\documentclass[10pt,twocolumn,letterpaper]{article}

\usepackage{cvpr}              

\usepackage[dvipsnames]{xcolor}

\usepackage[accsupp]{axessibility}
\usepackage{graphicx}
\usepackage{amsmath}
\usepackage{amssymb}
\usepackage[dvipsnames]{xcolor}
\usepackage{amsfonts}
\usepackage{bbm}
\usepackage{algorithm}
\usepackage{listings}
\usepackage{makecell} 
\usepackage{booktabs} 
\usepackage{colortbl}
\usepackage{multirow}
\usepackage{balance} 
\usepackage{pifont}
\definecolor{cvprblue}{rgb}{0.21,0.49,0.74}
\usepackage[pagebackref,breaklinks,colorlinks,citecolor=cvprblue]{hyperref}
\definecolor{codegreen}{rgb}{0,0.6,0}
\definecolor{codegray}{rgb}{0.5,0.5,0.5}
\definecolor{graphgreen}{RGB}{33, 140, 33}
\definecolor{resultgreen}{RGB}{0, 166, 79}
\definecolor{resultbg}{RGB}{230, 247, 254}
\definecolor{resultbgsim}{RGB}{200, 235, 235}
\lstdefinestyle{mystyle}{
    commentstyle=\color{codegreen},
    keywordstyle=\color{magenta},
    numberstyle=\tiny\color{codegray},
    stringstyle=\color{codepurple},
    basicstyle=\ttfamily\footnotesize,
    breaklines=true,                 
    keepspaces=true,                 
    numbers=left,                    
    showstringspaces=false,
}
\def\paperID{3357} 
\def\confName{CVPR}
\def\confYear{2024}

\title{Solving the Catastrophic Forgetting Problem in Generalized Category Discovery}

\author{Xinzi Cao\textsuperscript{1,2}\thanks{Equal Contribution.}, Xiawu Zheng\textsuperscript{2,3}$^{*}$, Guanhong Wang\textsuperscript{4}\\
Weijiang Yu\textsuperscript{1}, Yunhang Shen\textsuperscript{6}, Ke Li\textsuperscript{6}, Yutong Lu\textsuperscript{1}\thanks{Joint Corresponding Authors.}, Yonghong Tian\textsuperscript{2,8\textdagger}\\
[1mm]
\textsuperscript{1} Sun Yat-sen University,
\textsuperscript{2} Peng Cheng Laboratory,
\textsuperscript{3} Xiamen University\\
\textsuperscript{4} Zhejiang University, 
\textsuperscript{5} Tencent Youtu Lab,
\textsuperscript{6} Peking University\\
[1mm]
{\tt\small caoxz@mail2.sysu.edu.cn \quad zhengxiawu@xmu.edu.cn \quad guanhongwang@zju.edu.cn}\\ 
{\tt\small \{weijiangyu8, shenyunhang01\}@gmail.com \quad tristanli@tencent.com}\\
{\tt\small luyutong@mail.sysu.edu.cn \quad yhtian@pku.edu.cn}}

\begin{document}
\maketitle
\begin{abstract}
Generalized Category Discovery~(GCD) aims to identify a mix of known and novel categories within unlabeled data sets, providing a more realistic setting for image recognition.
Essentially, GCD needs to \textbf{remember} existing patterns thoroughly to recognize novel categories.
Recent state-of-the-art method SimGCD transfers the knowledge from known-class data to the learning of novel classes through debiased learning. 
However, some patterns are catastrophically \textbf{forgot} during adaptation and thus lead to poor performance in novel categories classification.
To address this issue, we propose a novel learning approach, \textbf{LegoGCD}, which is seamlessly integrated into previous methods to enhance the discrimination of novel classes while maintaining performance on previously encountered known classes.
Specifically, we design two types of techniques termed as \textbf{\underline{L}}ocal \textbf{\underline{E}}ntropy Re\textbf{\underline{g}}ularization~(LER) and Dual-views Kullback–Leibler divergence c\textbf{\underline{o}}nstraint~(DKL).
The LER optimizes the distribution of potential known class samples in unlabeled data, thus ensuring the preservation of knowledge related to known categories while learning novel classes.
Meanwhile, DKL introduces Kullback–Leibler divergence to encourage the model to produce a similar prediction distribution of two view samples from the same image.
In this way, it successfully avoids mismatched prediction and generates more reliable potential known class samples simultaneously. 
Extensive experiments validate that the proposed LegoGCD effectively addresses the known category forgetting issue across all datasets, \eg, delivering a $\textbf{7.74\%}$ and $\textbf{2.51\%}$ accuracy boost on known and novel classes in CUB, respectively. Our code is available at: 
\url{https://github.com/Cliffia123/LegoGCD}.
\end{abstract}    
\section{Introduction}
\begin{figure}
  \centering
  \begin{subfigure}{0.49\linewidth}
   \includegraphics[width=1\linewidth]{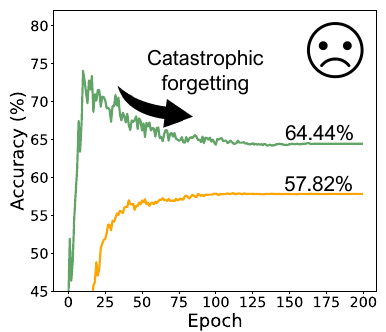}
    \caption{Baseline SimGCD \cite{wen2023parametric}}
    \label{fig:SimGCD}
  \end{subfigure}
  \begin{subfigure}{0.49\linewidth}
    \includegraphics[width=1\linewidth]{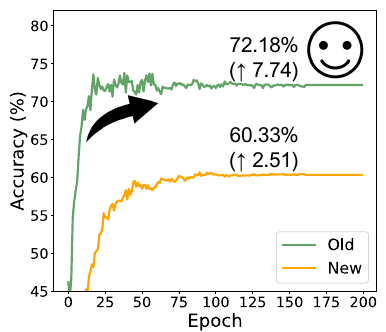}
    \caption{Ours LegoGCD}
    \label{fig:LegoGCD}
  \end{subfigure}
  \caption{Visualization of the accuracy results in unlabeled dataset on CUB dataset \cite{wah2011caltech} during training.
(a) shows a decrease in the accuracy of known (Old) classes ({\color{graphgreen}green}) in the baseline as the accuracy of novel (New) classes ({\color{orange}orange}) increases.
(b) demonstrates that LegoGCD solves the catastrophic forgetting problem and surpasses the baseline by a significant margin of $7.74$.}
  \label{fig:overview_compare}
\end{figure}

Deep learning have achieved superior performance on computer vision tasks~\cite{HeGDG17, RenHGS15, SzegedyLJSRAEVR15,DRonnebergerFB15, VaswaniSPUJGKP17,CaronTMJMBJ21}, particularly on image classification~\cite{HeZRS16, HuangLMW17, simonyan2014very, SandlerHZZC18, ZophVSL18}.
However, conventional methods work in a close-world setting, where all training data comes with pre-defined classes.
Consequently, deploying these models in real-world scenarios with potential novel classes becomes a considerable challenge.
Furthermore, these achievements rely heavily on large-scale annotated dataset, which is not easily accessible in realistic scenarios.
To address these challenges, a new paradigm of \textit{Generalized Category Discovery}~(GCD)~\cite{VazeHVZ22, ZhangKSNCK23, PuZS23, an2023generalized, wen2023parametric, ZhangMG023, FiniSLZN021, HanREVZ22} has been proposed and attracts increasing attention in recent years.

The goal of GCD is to train a classification model capable of recognizing both known and novel categories within unlabeled data.
To be clear, GCD distinguishes itself from the Novel Class Discovery~(NCD)~\cite{HanVZ19}, which relies on an unrealistic assumption that all unlabeled data exclusively belongs to entirely new classes or patterns.
In contrast, GCD adopts a more pragmatic assumption, acknowledging that unlabeled data encompasses a mixture of both known and novel categories. 
Consequently, GCD is more realistic compared to NCD, especially in real-world scenarios.

Since GCD is partially based on the learned patterns, an intuitive idea is to classify the unlabeled data through a clustering-based approach \cite{VazeHVZ22} \ie \textit{k}-means. 
However, as the scale of datasets increases, the computational costs for clustering in the original GCD grow exponentially.
To tackle this issue, Wen \etal introduce SimGCD \cite{wen2023parametric}, which replaces the clustering approach with a classifier.
Specifically, SimGCD trains the classifier using a pseudo-labeling strategy, a technique that has demonstrated remarkable effectiveness in Semi-supervised Learning~(SSL)~\cite{BerthelotCGPOR19, ZhangMG023}. 
Nevertheless, the pseudo labels of novel samples tend to be assigned as known classes due to the absence of guidance for novel class samples. 
In response, Wen \etal further propose to adopt class mean entropy to encourage the model to focus on novel categories, consequently generating high-quality pseudo labels for classifier training.  
As a result, SimGCD has achieved state-of-the-art performance and established itself as a robust baseline solution in the GCD setting.

However, SimGCD \cite{wen2023parametric} still has a significant drawback. It encourages the model to focus more on novel classes by employing an entropy regularization, which unfortunately comes at the cost of known class accuracy, resulting in a \textbf{catastrophic forgetting problem} in known categories. To illustrate this issue, we have tracked the classification accuracy of known and novel categories on unlabeled data during each training epoch on CUB \cite{wah2011caltech}. As shown in \cref{fig:SimGCD}, the {\color{graphgreen}green} curve represents the accuracy of known~(Old)~classes, while the {\color{orange}orange} curve represents the accuracy of novel~(New)~classes. Notably, we can easily observe this phenomenon, with the accuracy of novel classes improving, the accuracy of known categories initially increases to approximately 74\% after 20 epochs but then drops to 64.44\% in the end. We thus conclude that SimGCD faces catastrophic forgetting in known categories during training.

To address the above issue, we propose a novel \textit{Local Entropy Regularization}~(LER)~to preserve the knowledge of known categories. In particular, we first identify potential known samples using a threshold on their logits prediction like FixMatch \cite{SohnBCZZRCKL20}. Then, we employ the information entropy function to encourage the predictions of above selected known samples close to a more certain distribution, thereby increasing the confidence of samples from potential known classes. Consequently, this LER prevents known samples from being misclassified as novel classes and therefore preserves the knowledge related to known categories during learning novel categories.

It's worth noting that the model may occasionally miss potential known samples or select incorrect (novel) samples for LER. For example, when we have two augmentation view samples, $\boldsymbol{x}_i$ and $\boldsymbol{x}^{\prime}_{i}$, from the same image, and $\boldsymbol{x}_i$ has higher logits than the threshold while $\boldsymbol{x}^{\prime}_{i}$ falls below it. In such cases, we can't be certain whether the original image belongs to known classes, and this uncertainty may impact the effectiveness of LER. We argue that the predictions of the two view samples should be correctly aligned to ensure the quality of the chosen known samples. Therefore, we further propose a dual-view alignment scheme called \textit{Dual-views Kullback–Leibler divergence constraint}~(DKL), which employs Kullback–Leibler (KL) divergence to encourage the consistency of two views from the same image.

To summarize, we propose a novel approach named LegoGCD, which integrates SimGCD \cite{wen2023parametric} with our proposed LER and DKL to address the problem of catastrophic forgetting. To validate the effectiveness of LegoGCD, we conduct extensive experiments on eight datasets, including generic datasets such as CIFAR10/100 \cite{krizhevsky2009learning}, ImageNet-1k \cite{DengDSLL009}, and fine-grained datasets CUB \cite{wah2011caltech}, Stanford Cars \cite{Krause0DF13}, and FGVC-Aircraft \cite{MajiRKBV13}. Intuitively, we also visualize the classification accuracy of known and novel categories on CUB \cite{wah2011caltech}. These results are shown in \cref{fig:LegoGCD}. The {\color{graphgreen} green}  curve represents the accuracy of known (Old) categories, while the {\color{orange}orange} curve indicates the accuracy of novel (New) classes. Clearly, LegoGCD effectively prevents the decline in known classes and achieves an accuracy of 72.18\%, surpassing SimGCD by a margin of \textbf{7.74}. Clearly, the results indicate that LegoGCD solves the catastrophic forgetting problem of known categories effectively. Moreover, our method can be easily placed onto SimGCD like Lego, requiring only a few lines of code on the implementation without introducing any additional parameters or altering the internal network structure of SimGCD.

In summary, our key contributions are as follows:
\begin{itemize}
\item We introduce a novel constraint named Local Entropy Regularization~(LER), which is designed to mitigate the catastrophic forgetting problem of known classes by preserving the knowledge of known categories during learning novel classes.


\item We propose a Dual-views Kullback–Leibler divergence constraint~(DKL)~that ensures the prediction distribution of one view approximates that of another,  maintaining consistency between dual views augmented from the same image.



\item The proposed LegoGCD is effective and can be simply integrated from SimGCD without any extra parameter addition. Extensive results demonstrate our method exhibits significant performance improvement on known classes, \eg, a 7.74\% increase in CUB \cite{wah2011caltech}.
\end{itemize}
\section{Related Work}
\subsection{Generalized Category Discovery}GCD was first formulated by Vaze \etal \cite{VazeHVZ22}, presents a unique challenge distinct from Semi-supervised Learning~(SSL)~\cite{BerthelotCGPOR19, MiyatoMKI19, ZhangZHZMCO22, TarvainenV17, LaineA17}. While SSL assumes that unlabeled data belongs to the same class set as the labeled data, GCD tackles a more realistic scenario where the unlabeled data may include classes not present in the labeled set, which is the same setting in Novel Category Discovery (NCD) \cite{HanVZ19, HanREVZ22, FiniSLZN021, JosephPABRHB22, ZhaoZSL22, YangZYWD22, LiFHG23, YangWDZ23, LiuWZFYS23}. 
Therefore,  GCD can be viewed as an extension of NCD, with the main difference being that GCD seeks to identify specific categories within novel classes, while NCD focuses on grouping novel classes into a single category.
The original GCD approach employs contrastive and SSL, which uses clustering during inference and incurs significant computational costs. To address this challenge, Wen \etal \cite{wen2023parametric} introduce SimGCD with a classifier to replace clustering, offering a robust baseline for the GCD problem. However, it's important to note that SimGCD introduces a drawback, leading to a decrease in the classification accuracy of known classes during the intense learning of novel categories.

\subsection{Entropy regularization} It is a widely used technique in image classification, especially in the context of cross-entropy, which aims to align prediction distributions with the standard label distribution. However, in scenarios such as Semi-supervised Learning~(SSL)~\cite{BerthelotCGPOR19, MiyatoMKI19, ZhangZHZMCO22, TarvainenV17, LaineA17}, where true labels are unknown, pseudo labels take the place of actual labels in standard cross-entropy. This form of entropy regularization minimizes output differences between various views of unlabeled data. Notably, Data Augmentations \cite{YunHCOYC19, ZhangCDL18} have proven effective, contributing to substantial successes in pseudo-supervised learning. For instance, in SimGCD \cite{wen2023parametric}, an augmentation strategy generates two views of data, establishing training targets in one view and enforcing prediction consistency with the other view during unlabeled data training. Another form of entropy, information entropy, measures the amount of information within a set of events. In SimGCD, information entropy is employed to minimize class mean entropy, promoting more uniform class predictions in each iteration to ensure the visibility of novel classes. However, due to the absence of protection for the knowledge of known classes, class mean entropy has led to a degeneration in known categories.
\section{Method}
In this section, we first formulate the GCD task (\cref{formulation}) and present the overview of the baseline SimGCD \cite{wen2023parametric} (\cref{preliminaries}). Then, we introduce how to mitigate the degradation of SimGCD by the proposed LegoGCD. At last, we describe the details of the proposed Local Entropy Regularization~(LER)~and Dual-views Kullback-Leibler divergence constraint~(DKL)~in \cref{LER} and \cref{DKL}, respectively.

\begin{figure*}[t]
	\centering
	{\includegraphics[width=1\linewidth]{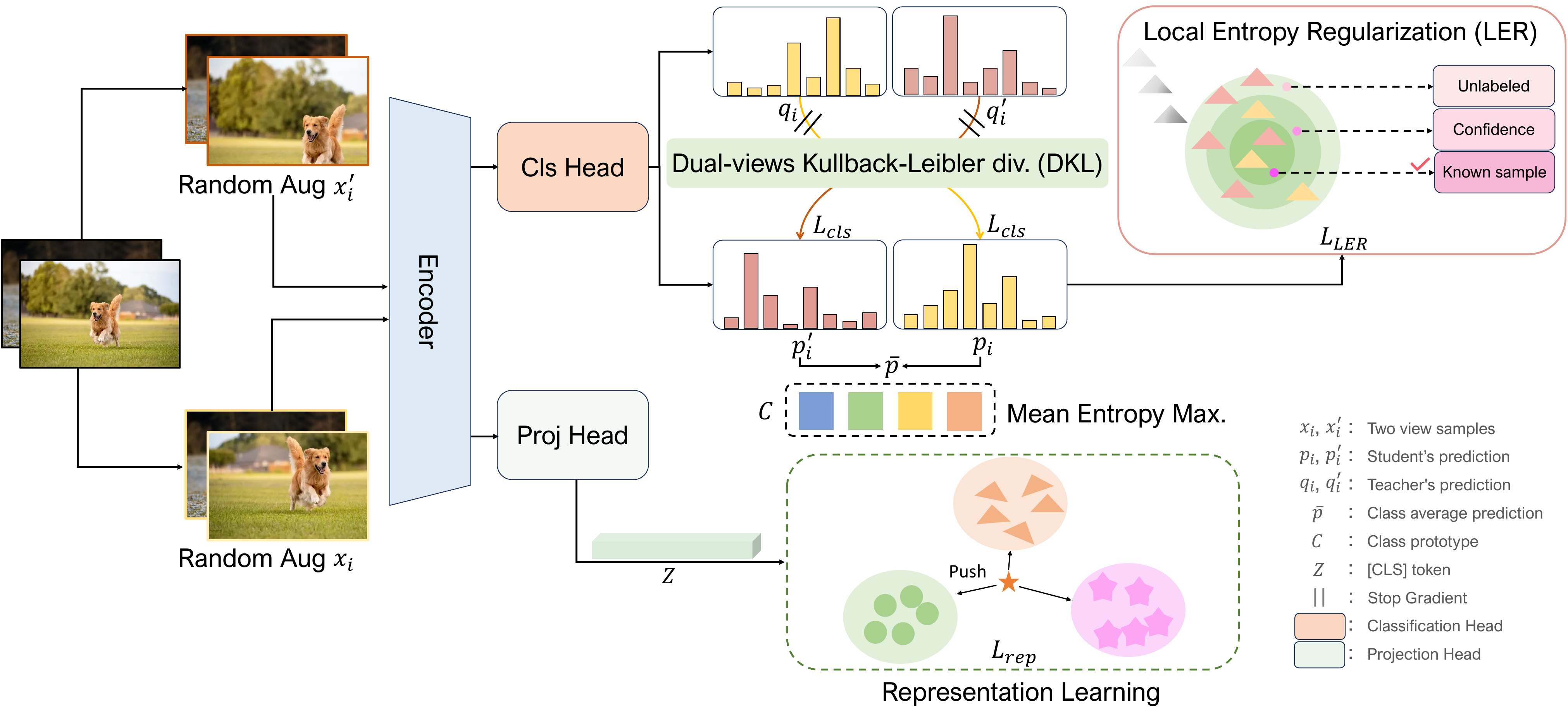}}
	\caption{Illustration of our proposed LegoGCD. LegoGCD is mainly composed of SimGCD and our proposed LER and DKL. (a) Representation learning and Mean Entropy in SimGCD (\cref{preliminaries}). (b) Local Entropy Regularization (LER) (\cref{LER}) for discovering potential known samples in unlabeled data and preserving the knowledge of known classes. (c) Dual-views Kullback-Leibler divergence (DKL) (\cref{DKL}) to ensure consistent predictions for two view samples.}
	\label{fig:framework}
\end{figure*}

\subsection{Problem Formulation}{\label{formulation}}
Traditional image classification tasks are typically developed using a labeled dataset, denoted as $\mathcal{D}^l\!=\!\{(\boldsymbol{x}_i, y_i)\}\!\in\mathcal{X}\!\times\!\mathcal{Y}_l$. This dataset contains only samples from known classes, represented by $\mathcal{Y}_l$. In contrast, Generalized Category Discovery~(GCD)~aims to recognize unlabeled data, denoted as $\mathcal{D}^u\!=\!\{(\boldsymbol{x}_i, y_i)\}\!\in\!\mathcal{X}\!\times\!\mathcal{Y}_u$. This dataset comprises both known and novel class samples, where $\mathcal{Y}_l$ is a subset of $\mathcal{Y}_u$. The goal of GCD is to develop a model that can identify both known and novel classes using the labels from known categories~($\mathcal{Y}_l$)~and unlabeled data~($\mathcal{D}_u$)~without access to class labels. It's important to note that the total number of categories is represented as $K=\left|\mathcal{Y}_l \cup\!\mathcal{Y}_u\right|$. We assume prior knowledge of this total category count, as done in previous works \cite{FiniSLZN021, HanREVZ22, ZhaoH21, ZhongZLL0S21}.
\subsection{Preliminaries}{\label{preliminaries}}
In this section, we introduce the fundamental structure of SimGCD \cite{wen2023parametric} baseline program as shown in \cref{fig:framework}. This program primarily consists of two key components: Representation learning and Parametric Classification.

\subsubsection{Representation Learning}
The representation learning process in our framework follows GCD \cite{VazeHVZ22} and SimGCD \cite{wen2023parametric}. It employs a Vision Transformer~(ViT-B/16)~\cite{DosovitskiyB0WZ21} pretrained with DINO self-supervision \cite{CaronTMJMBJ21} on ImageNet \cite{DengDSLL009} as the backbone. This process includes supervised contrastive learning on labeled data and unsupervised contrastive learning on \textit{all} data, encompassing both labeled and unlabeled data.

Formally, given two views~(random augmentations)~$\boldsymbol{x}_i$ and $\boldsymbol{x}_i^{\prime}$ of the same image in a mini-batch $B$, the unsupervised contrastive loss is written as:
\begin{equation}\label{unsup_con_loss}
\mathcal{L}_{\text {rep }}^u=\frac{1}{|B|} \sum_{i \in B}-\log \frac{\exp \left(\boldsymbol{z}_i^{\top} \cdot \boldsymbol{z}_i^{\prime} / \tau_u\right)}{\sum_i^{i \neq n} \exp \left(\boldsymbol{z}_i^{\top} \cdot \boldsymbol{z}_n^{\prime} / \tau_u\right)},
\end{equation}
where $\boldsymbol{z}_i=\phi\left(f\left(\boldsymbol{x}_i\right)\right)$ and $\boldsymbol{z}_i^{\prime} =\phi\left(f\left(\boldsymbol{x}_i^{\prime}\right)\right)$, $f$ is the feature backbone, $\phi$ is a multi-layer perceptron~(MLP)~projection head, $\tau_u$ is a unsupervised temperature.

The objective of the supervised contrastive loss is to encourage the model to bring samples with the same class label closer in the feature space, formally written as:
\begin{equation}\label{sup_con_loss}
\mathcal{L}_{\text {rep }}^s\!=\!\frac{1}{\left|B^l\right|}\!\!\sum_{i \in B^l}\!\frac{1}{\left|\mathcal{N}_i\right|}\!\!\sum_{q \in N_i}\!-\!\log\!\frac{\exp\!\left(\boldsymbol{z}_i^{\top}\!\!\cdot\!\boldsymbol{z}_q^{\prime}/\tau_c\right)}{\sum_i^{i \neq n}\!\exp\!\left(\boldsymbol{z}_i^{\top}\!\!\cdot\!\boldsymbol{z}_n^{\prime}/\tau_c\right)},
\end{equation}
where $\mathcal{N}_i$ is the indices of images share the same label with $\boldsymbol{x}_i$ in the mini-batch ${B}$. Finally, the total loss in representation learning is constructed as: 
\begin{equation}
\mathcal{L}_{\text {rep }}=(1-\lambda) \mathcal{L}_{\text {rep }}^u+\lambda \mathcal{L}_{\text {rep }}^s,
\end{equation}
where $B^l$ is the labeled subset of $B$ and $\lambda$ is a weight factor.

\subsubsection{Parametric Classification}\label{parametric_classification}
Different from the GCD \cite{VazeHVZ22} that uses $k$-means, SimGCD \cite{wen2023parametric} employs a more efficient classifier based on self-distillation \cite{CaronTMJMBJ21, AssranCMBBVJRB22}. Formally, the classifier is denoted as a set of prototypes $\mathcal{C}=\left\{\boldsymbol{c}_1, \ldots, \boldsymbol{c}_K\right\}$, where  $K=\left|\mathcal{Y}_l \cup \mathcal{Y}_u\right|$.  During training, the soft label of each view  $\boldsymbol{x}_i$ is obtained by applying softmax on cosine similarity between hidden feature $\boldsymbol{h}_i =f\left(\boldsymbol{x}_i\right)$ and prototypes $\mathcal{C}$, scaled by $1/\tau_s$:
\begin{equation}
\boldsymbol{p}_i^{(k)}=\frac{\exp \left(\frac{1}{\tau_s}\left(\boldsymbol{h}_i /\left\|\boldsymbol{h}_i\right\|_2\right)^{\top}\left(\boldsymbol{c}_k /\left\|\boldsymbol{c}_k\right\|_2\right)\right)}{\sum_{k^{\prime}} \exp \left(\frac{1}{\tau_s}\left(\boldsymbol{h}_i /\left\|\boldsymbol{h}_i\right\|_2\right)^{\top}\left(\boldsymbol{c}_{k^{\prime}} /\left\|\boldsymbol{c}_{k^{\prime}}\right\|_2\right)\right)},
\end{equation}
The soft pseudo label $\boldsymbol{q}_i^{\prime}$ for view $\boldsymbol{x}_i^{\prime}$ is similarly generated. Then, it employs a cross-entropy loss $\ell\left(\boldsymbol{q}^{\prime}, \boldsymbol{p}\right)=-\sum_k \boldsymbol{q}^{\prime(k)} \log \boldsymbol{p}^{(k)}$ to supervise the learning of prediction with pseudo labels or ground-truth labels:
\begin{equation}{\label{loss_cls}}
\mathcal{L}_{\mathrm{cls}}^{u}=\frac{1}{|B|} \sum_{i \in B} \ell\left(\boldsymbol{q}_i^{\prime}, \boldsymbol{p}_i\right), \mathcal{L}_{\mathrm{cls}}^s=\frac{1}{|B|} \sum_{i \in B} \ell\left(\boldsymbol{y}_i, \boldsymbol{p}_i\right),
\end{equation}
where $\mathcal{L}_{\mathrm{cls}}^{u}$ and $\mathcal{L}_{\mathrm{cls}}^s$ are unsupervised and supervised classification losses for {all} and labeled data, respectively. 
To regulate unsupervised learning, SimGCD adopts a class mean entropy regulariser \cite{AssranCMBBVJRB22}: $H(\overline{\boldsymbol{p}})=-\sum_k \overline{\boldsymbol{p}}^{(k)} \log \overline{\boldsymbol{p}}^{(k)}$, where $\overline{\boldsymbol{p}}$ is the mean prediction of each class in a batch $\overline{\boldsymbol{p}}=\frac{1}{2|B|} \sum_{i \in B}\left(\boldsymbol{p}_i+\boldsymbol{p}_i^{\prime}\right)$. Then the classification objective is: $\mathcal{L}_{\mathrm{cls}}=(1-\lambda)( \mathcal{L}_{\mathrm{cls}}^u-\varepsilon H(\overline{\boldsymbol{p}}))+\lambda \mathcal{L}_{\mathrm{cls}}^s
$. Finally, the overall objective in baseline SimGCD is $\mathcal{L}_{\mathrm{rep}}+\mathcal{L}_{\mathrm{cls}}$.

\subsection{Local Entropy Regularization}\label{LER}
\begin{figure}
  \centering
  \begin{subfigure}{0.49\linewidth}
   \includegraphics[width=1\linewidth]{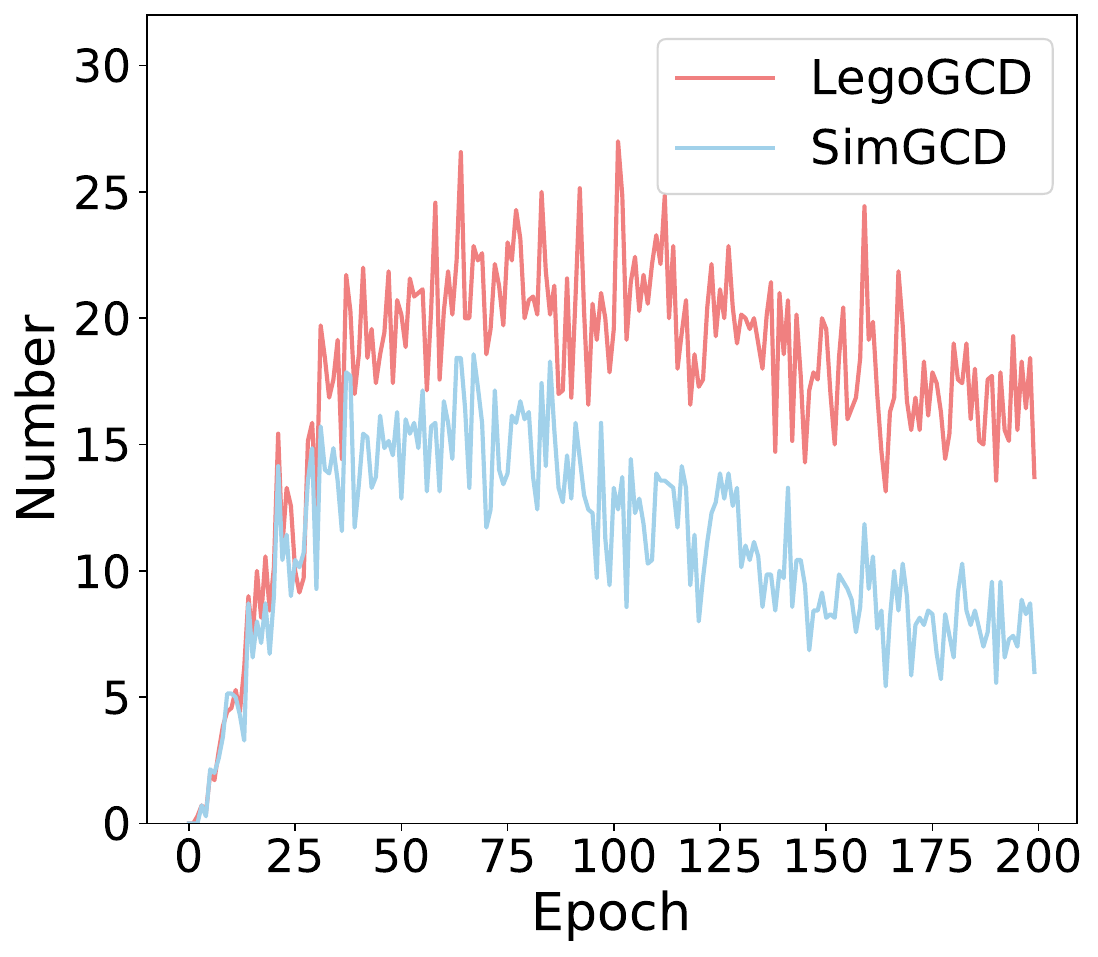}
    \caption{FGVC-Aircraft \cite{MajiRKBV13}}
    \label{fig:FGVC-Aircraft}
  \end{subfigure}
  \begin{subfigure}{0.5\linewidth}
    \includegraphics[width=1\linewidth]{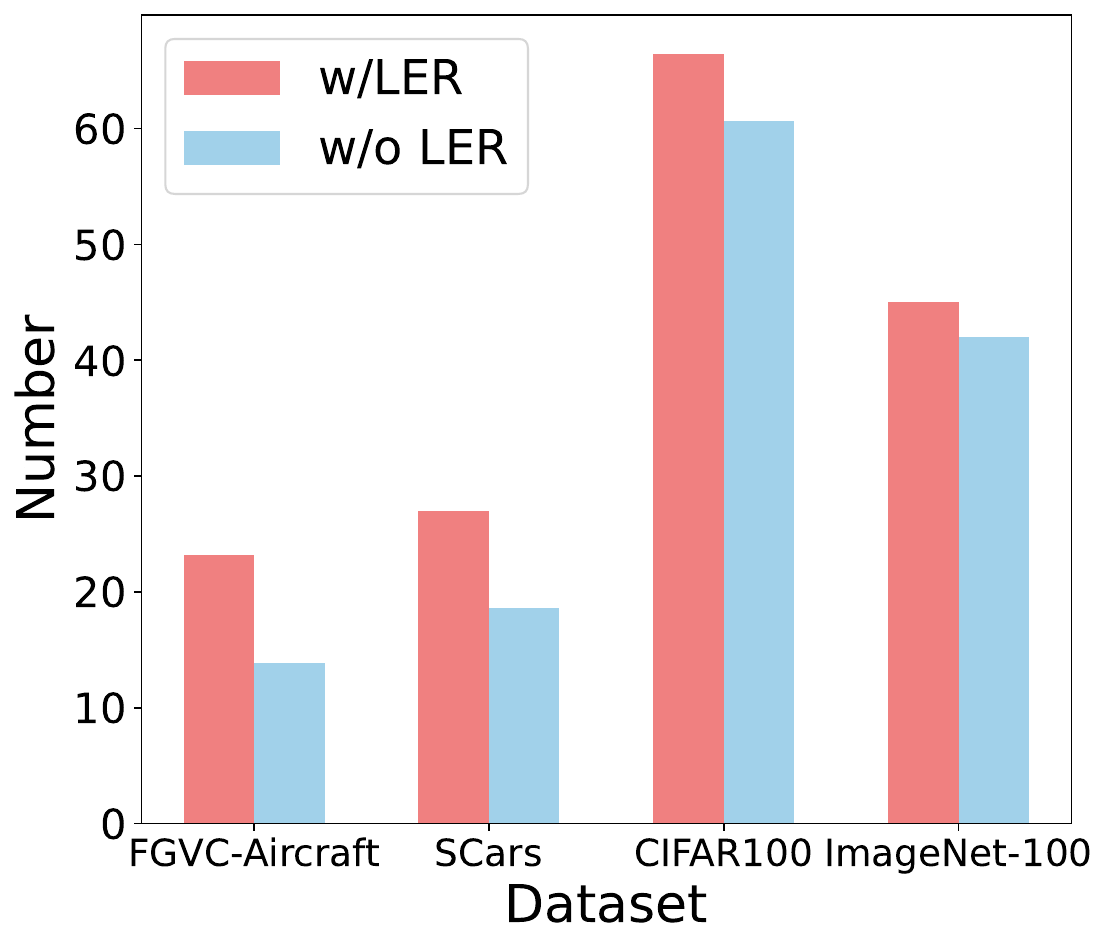}
    \caption{Maximum numbers}
    \label{fig:number_in_all_datasets}
  \end{subfigure}
\caption{Comparison of potential known samples in SimGCD \cite{wen2023parametric} and LegoGCD. (a) LegoGCD recognizes almost 10 more high-confidence samples than SimGCD in the end on the FGVC-Aircraft dataset. (b) LegoGCD with LER produces more high-confidence known samples in various generic and fine-grained datasets compared to SimGCD without LER.}
	\label{fig:potential_known_samples}
\end{figure}

\textbf{Motivation.}
Although the baseline SimGCD \cite{wen2023parametric} has improved accuracy and computational efficiency compared to GCD \cite{VazeHVZ22}, it faces catastrophic forgetting in known (Old) classes during training, as shown in \cref{fig:SimGCD}. This is due to SimGCD relying on class mean entropy, causing a shift in focus to novel classes and resulting in information loss in known classes. We argue that retaining knowledge of known classes should be prioritized. \cref{fig:FGVC-Aircraft} displays potential known class samples in each epoch on the FGVC-Aircraft \cite{MajiRKBV13} dataset. Evidently, LegoGCD recognizes nearly 10 more potential known samples than SimGCD in the end. \cref{fig:number_in_all_datasets} illustrates the maximum number of potential known samples in various datasets, confirming that LegoGCD with LER excels at preserving the recognition ability of known categories.
Concretely, we propose a Local Entropy Regularization~(LER)~to preserve the knowledge of known categories, solving the problem of catastrophic forgetting. In contrast to the class mean entropy, which primarily shifts the network's focus to novel classes, we argue that the network should also maintain its ability to recognize known samples as it did before. Specifically, we choose known samples based on a confidence threshold $\delta$ in unlabeled data and utilize entropy regularization to ensure the stability of these known classes associated with the selected known samples.

The training dataset, denoted as $\mathcal{D}\!=\!\{(\boldsymbol{x}_i, y_i)\}\!\in\!\mathcal{X}\!\times\!\mathcal{Y}$, comprises both labeled and unlabeled samples represented as $\boldsymbol{x}_i$ within a batch $B$. To distinguish between labeled and unlabeled samples in a batch, we utilize a binary mask vector ${M}=[m_1, m_2, \ldots, m_i] \subseteq\{0,1\}$, where each $m_i$ can be either 0~(indicating an unlabeled sample)~or 1~(indicating a labeled sample). Consequently, the unlabeled samples in a batch are obtained by applying this mask $M=0$.

Next, let $\boldsymbol{p}_{i}=\left[p^1_{i}, p^2_{i}, \ldots, p^k_{i}\right]$ as the prediction vector for sample $\boldsymbol{x}_i$, where ${K}$ represents the total number of categories. We use $S=\left[s_1, s_2, \ldots, s_i\right]\subseteq\{0,1\}$ as a binary vector to denote the \textbf{high-confidence} sample. When $s_{i} = 1$, it indicates that $\boldsymbol{x}_i$ is a potential high-confidence sample. This can be expressed as follows:
\begin{equation}
s_{i}=\mathbbm{1}\left(\max \left(\boldsymbol{p}_{i}\right) \geq \delta \right),
\end{equation}
where $\delta$ represents the confidence threshold. Then, we let $\mathcal{Y}=\left[y_1, y_2, \ldots, y_i\right] \in \{1, 2, \ldots, K\}$ denotes the potential class label corresponding to $\boldsymbol{x}_i$. $y_i$ is determined by the index of the maximum value in $\boldsymbol{p}_{i}$:
\begin{equation}
y_i = \arg\max_j \left( \boldsymbol{p}_{i} \right)_j, \quad i = 1, 2, \ldots, b
\end{equation}
where $b$ denotes the number of batch sizes. We also introduce  $\mathcal{O}=\left[o_1, o_2, \ldots, o_i\right] \subseteq\{0,1\}$ as a binary vector, and $o_{i}=1$ signifying potential \textbf{known} samples with \textbf{high-confidence} in \textbf{unlabeled data}. This calculation can be performed as follows:
\begin{equation}
 o_{i} =  \underbrace{\mathbbm{1}\left(m_{i}=0\right)}_{\text{Unlabeled}} \cdot \underbrace{\mathbbm{1}\left(s_{i}=1\right)}_{\text{High-confidence}} \cdot \underbrace{\mathbbm{1}\left({y}_i \in \mathcal{Y}_l \right)}_{\text{Known}},
\end{equation}
where $ \mathcal{Y}_l $ represents the known class set. Next, we use information entropy to assess the stability of the known categories during training, which is expressed as follows:
\begin{equation}\label{loss_ler_entropy}
\mathcal{L}_{entropy} = - \frac{1}{B} \sum_{i=1}^{B} \mathbbm{1}\left(o_i=1\right) \cdot {H}\left(\frac{1}{\tau_{o}}{\boldsymbol{p}}(\boldsymbol{x}_i)\right),
\end{equation}
where ${\tau_{o}}$ is a temperature, $H$ is an entropy resulariser \cite{AssranCMBBVJRB22} used in \cref{parametric_classification}. Additionally, to further enhance the margins between the categories, we replace the vanilla information loss with a Margin-aware Pattern (MAP) \cite{WangWLY22, YuLL23}, and the final LER loss can be formulated as:
\begin{equation}{\label{MAP}}
\!\!\mathcal{L}_{LER}\!=\!\frac{1}{B}\!\sum_{i=1}^{B}\!\mathbbm{1}\!\left(o_i\!=\!\!1\right)\cdot{H}\!\left(\!\frac{1}{\tau_{o}}{\boldsymbol{p}}\!\left({\boldsymbol{x}_i}\right),\!\frac{1}{\tau_{o}}{\boldsymbol{p}}\!\left({\boldsymbol{x}_i}\right)\!+\!{\Delta_j}\right),
\end{equation}
where ${\Delta_j}=\lambda_{ler} \log \left(\frac{1}{\tilde{p}_j}\right)$, $j \in\{1, \ldots, K\}$, $\lambda_{ler}=0.4$, and $\tilde{p}_j$ is the average model prediction updated at each iteration through an exponential moving average.

The total process is divided into three steps: 1) Applying the label mask $M$ to select samples from \textbf{unlabeled data}; 2) Using the threshold $\delta$ to select \textbf{high-confidence} samples from all training data; 3) Identifying high-confidence \textbf{known} samples from all training samples. Furthermore, we provide an intuitive representation of the entire process in the top-right corner of \cref{fig:framework}. 

\cref{fig:number_in_all_datasets} compares the quantity of high-confidence known samples across various datasets. Clearly, the numbers increase with the introduction of LER, proving our method with LER retains more knowledge about known classes.


\subsection{Dual-views Kullback-Leibler divergence}{\label{DKL}}

It's crucial to note that the model might mistakenly identify incorrect samples as known ones when considering two views of the same image for LER. For instance, if $\boldsymbol{x}_i$ and $\boldsymbol{x}^{\prime}_i$ are misaligned, and $\boldsymbol{x}_i$ exceeds the confidence threshold $\delta$ while $\boldsymbol{x}^{\prime}_i$ falls below it, uncertainty arises about whether the image of the two views is a potential known one. Thus, we might erroneously select $\boldsymbol{x}_{i}$ or miss $\boldsymbol{x}^{\prime}_i$ for LER. In other words, we can confidently select both $\boldsymbol{x}_i$ and $\boldsymbol{x}_i^{\prime}$ as known samples only when both $\boldsymbol{x}_i$ and $\boldsymbol{x}_i^{\prime}$ exceed the threshold $\delta$.

In general, an ideal approach is to push the alignment of two view samples $\boldsymbol{x}_i$ and $\boldsymbol{x}_i^{\prime}$ and both belong to the known or novel sample set. To achieve this, we propose a dual-view alignment technique named Dual-views Kullback-Leibler divergence constraint~(DKL).

Formally, given wo cosine similarity  $\boldsymbol{p}_i$ and $\boldsymbol{p}_i^{\prime}$ from \cref{parametric_classification} of two view samples $\boldsymbol{x}_i$ and $\boldsymbol{x}^{\prime}_i$ in a mini-batch $B$, the DKL can be formulated as: 
\begin{equation}
D_{KL}(\boldsymbol{p}_i \| \boldsymbol{p}_i^{\prime})=\frac{1}{B/2} \sum_{i=1}^{B/2} \boldsymbol{p}\left(\boldsymbol{x}_i\right) \cdot \log \frac{\boldsymbol{p}  \left(\boldsymbol{x}_i\right)}{\boldsymbol{p}\left(\boldsymbol{x}_i^{\prime}\right)}.
\end{equation}

In summary, DKL aligns the predictions of two view samples, enhancing the creation of more reliable potential known samples for LER. Finally, the ultimate classification loss can be updated as:
\begin{equation}
\mathcal{L}_{\mathrm{cls}}=(1-\lambda)( \mathcal{L}_{\mathrm{cls}}^u-\varepsilon H(\overline{\boldsymbol{p}})+D_{KL}(\boldsymbol{p}_i \| \boldsymbol{p}_i^{\prime}))+\lambda \mathcal{L}_{\mathrm{cls}}^s,
\end{equation}
where $\lambda$ is a weight factor to control the balance between supervised and unsupervised classification learning.

By simply integrating the Local Entropy Regularization (LER) and Dual-views Kullback-Leibler divergence constraint~(DKL)~into SimGCD, we propose a new paragram named LegoGCD. The overall loss for training our model can be formulated as:
\begin{equation}
\mathcal{L}=\alpha \cdot (\mathcal{L}_{rep}+\mathcal{L}_{cls})+\beta \cdot \mathcal{L}_{LER},
\end{equation}
where $\beta$ is a control factor to assign the weight to remember known classes. Aligning with SimGCD, we set $\alpha$ to 1 and simultaneously adjust $\beta$~(see \cref{tab:abs_different_thr}). Notably, the DKL is plugged into classification loss $\mathcal{L}_{cls}$. Algorithm 1 in appendix describes one training step of LegoGCD.

\section{Experiment}
\subsection{Experimental Setup}
\textbf{Dataset.} We evaluate the effectiveness of our approach on eight datasets, consistent with SimGCD \cite{wen2023parametric}. These datasets encompass generic image recognition datasets like CIFAR10/100 \cite{krizhevsky2009learning} and ImageNet-100 \cite{TianKI20}, as well as Semantic Shit \cite{Vaze0VZ22} datasets, including CUB \cite{wah2011caltech}, Stanford Cars \cite{Krause0DF13}, and FGVC-Aircraft \cite{MajiRKBV13}. Additionally, we include two more challenging datasets: Herbarium 19 \cite{tan2019herbarium} and ImageNet-1k \cite{RussakovskyDSKS15}. For each dataset, we follow the GCD \cite{VazeHVZ22} and SimGCD \cite{wen2023parametric} protocols by sub-sampling 50\% of known class images to form the labeled set $\mathcal{D}^l$ within the training set. The remaining images from known and novel classes constitute the unlabeled data $\mathcal{D}^u$. \cref{tab:dataset_show} provides details of the datasets used in our experiments.

\textbf{Evaluation protocol.} During training, we use dataset $\mathcal{D}$ combined by $\mathcal{D}^l$ and $\mathcal{D}^u$ to train the models.  For evaluation, we use clustering accuracy~(ACC)~\cite{VazeHVZ22, wen2023parametric} to evaluate the model performance. Specifically, ACC is computed as follows: given the ground-truth label $y^*$ and the model's prediction $\hat{y}_i$, $ACC=\frac{1}{M} \sum_{i=1}^M \mathbbm{1}\left(y_i^*=p\left(\hat{y}_i\right)\right)$, where $M=\left|\mathcal{D}^u\right|$, and $p$ is determined using the Hungarian optimal assignment algorithm \cite{kuhn1955hungarian}.

\begin{table}[t]
\caption{Overview of the datasets used in our experiments. We list the specific number of labeled and unlabeled images ($\mathcal{D}^l$, $\mathcal{D}^u$) and their corresponding class assignments (Old and New).}
    \centering
     \vspace{-3mm}
    \resizebox{\linewidth}{!}{
    \begin{tabular}{ccccccc}
    \toprule
         & &  \multicolumn{2}{c}{Labeled $\mathcal{D}^l$} & \multicolumn{2}{c}{Unlabeled $\mathcal{D}^u$}\\ \cmidrule(lr){3-4} \cmidrule(lr){5-6} 
         Dataset & Balance &  Images & Old & Images & New \\ \hline
         CIFAR10 \cite{krizhevsky2009learning}&\ding{51} &12.5k & 5 & 37.5k & 10\\
         CIFAR100 \cite{krizhevsky2009learning} &\ding{51} & 20.0k & 80 & 30.0k & 100\\
         ImageNet-100 \cite{TianKI20}&\ding{51} &31.9k & 50 & 95.3k & 100\\
         CUB \cite{wah2011caltech}&\ding{51} & 1.5k & 100 & 4.5k & 200\\
         Stanford Cars \cite{Krause0DF13}&\ding{51} & 2.0k & 98 & 6.1k &196 \\
         FGVC-Aircraft \cite{MajiRKBV13}&\ding{51} & 1.7k & 50 & 5.0k &100 \\
         Herbarium 19 \cite{tan2019herbarium}&\ding{55} & 8.9k & 341 & 25.4k & 683\\
         ImageNet-1K \cite{RussakovskyDSKS15}&\ding{51} & 321k & 500 & 960k &1000 \\
    \bottomrule
    \end{tabular}}
    \label{tab:dataset_show}
\end{table}

\begin{table*}[t]
    \caption{Classification results on Semantic Shift Benchmark \cite{Vaze0VZ22} datasets and Herbarium 19 \cite{tan2019herbarium}. \textbf{Bold}  represent our results, $\Delta$ indicates the margin ahead of the baseline SimGCD, and {\color{red}red} signifies improvement in known categories.}
    \vspace{-3mm}
    \centering
    \small
    \resizebox{0.95\linewidth}{!}{
    \begin{tabular}{ccccccccccccc}
    \toprule
&\multicolumn{3}{c}{CUB}&\multicolumn{3}{c}{Stanford Cars}&\multicolumn{3}{c}{FGVC-Aircraft} &\multicolumn{3}{c}{Herbarium 19}\\ \cmidrule(lr){2-4} \cmidrule(lr){5-7} \cmidrule(lr){8-10} \cmidrule(lr){11-13}
Method &All&Old&New &All&Old&New &All&Old& New&All& Old&New\\ \hline
\textit{k}-means\cite{macqueen1967some}& 34.3& 38.9 & 32.1 & 12.8 & 10.6 & 13.8 &  16.0& 14.4 & 16.8 & 13.0 &12.2 &13.4\\
RS+\cite{HanREVZ22} & 33.3 & 51.6 & 24.2 & 28.3 & 61.8 & 12.1 & 26.9 & 36.4 & 22.2 &27.9 &55.8 &12.8 \\
UNO+\cite{FiniSLZN021} &35.1 &49.0 &28.1 &35.5 &70.5 &18.6 &40.3 &56.4 &32.2 &28.3 &53.7 &14.7\\
ORCA\cite{LiuWZFYS23}  &35.3 &45.6 &30.2 &23.5 &50.1 &10.7 &22.0 &31.8 &17.1 &20.9 &30.9 &15.5\\ \hline
GCD \cite{VazeHVZ22} &51.3 &56.6 &48.7 &39.0 &57.6 &29.9 &45.0 &41.1 &46.9 &35.4 &51.0 &27.0\\
SimGCD \cite{wen2023parametric}&60.3 &65.6 &57.7 &53.8 &71.9 &45.0 &54.2 &59.1 &51.8 &44.0 &58.0 &36.4\\
\rowcolor{resultbgsim}
SimGCD$^{*}$ \cite{wen2023parametric}&61.9 &66.5 &59.6 &53.4	&70.6 &45.0 &54.6 &61.4 &51.1 &44.9 &56.9 &38.4\\ \hline
\rowcolor{resultbg}
Ours &\textbf{63.8}	&\textbf{71.9} &\textbf{59.8} &\textbf{57.3} &\textbf{75.7} &\textbf{48.4} &\textbf{55.0}  &\textbf{61.5} &\textbf{51.7} &\textbf{45.1} &\textbf{57.4} &\textbf{38.4} \\
\rowcolor{resultbg}
$\Delta$ &\textbf{1.9} &{\color{red}\textbf{5.4}} &\textbf{0.2} &\textbf{3.9} &{\color{red}\textbf{5.1}} &\textbf{3.4} &\textbf{0.4} &{\color{red}\textbf{0.1}} &\textbf{0.6} &\textbf{0.2} &{\color{red}\textbf{0.5}} & \textbf{0.0} \\
    \bottomrule
    \end{tabular}}
    \label{tab:result_legegcd_1}
\end{table*}

\begin{table}[t]
    \caption{Classification results on the generic image recognition datasets, CIFAR10 \cite{krizhevsky2009learning} and CIFAR100 \cite{krizhevsky2009learning}.} 
    \vspace{-3mm}
    \centering
    \resizebox{\linewidth}{!}{
    \begin{tabular}{ccccccc}
    \toprule
       &\multicolumn{3}{c}{CIFAR10}&\multicolumn{3}{c}{CIFAR100} \\ \cmidrule(lr){2-4} \cmidrule(lr){5-7}
        Method &All&Old&New &All&Old&New \\ \hline
        \textit{k}-means\cite{macqueen1967some} &83.6 &85.7 &82.5 &52.0 &52.2 &50.8 \\
        RS+\cite{HanREVZ22} &  46.8 &19.2 &60.5 &58.2 &77.6 &19.3  \\
        UNO+\cite{FiniSLZN021} &  68.6 &98.3 &53.8 &69.5 &80.6 &47.2  \\
        ORCA\cite{LiuWZFYS23} &  81.8 &86.2 &79.6 &69.0 &77.4 &52.0 \\ \hline
        GCD \cite{VazeHVZ22} & 91.5 &97.9 &88.2 &73.0 &76.2 &66.5 \\
        SimGCD \cite{wen2023parametric}& 97.1 &95.1 &98.1 &80.1 &81.2 &77.8 \\
        \rowcolor{resultbgsim}
        SimGCD$^{*}$ \cite{wen2023parametric}& 96.9 &93.8 &98.5&79.0 &77.9 &81.5\\ \hline
        \rowcolor{resultbg}
        Ours &\textbf{97.1} & \textbf{94.3} & \textbf{98.5} &\textbf{81.8} &\textbf{81.4}	&\textbf{82.5} \\  
        \rowcolor{resultbg}
       $\Delta$ &\textbf{0.2} &{\color{red}\textbf{0.5}} &\textbf{0.0} &\textbf{2.8} &{\color{red}\textbf{3.5}} &\textbf{1.0}  \\
    \bottomrule
    \end{tabular}}
    \label{tab:result_legegcd_CIFAR}
\end{table}

\begin{table}[t]
    \caption{Classification results on the generic image recognition ImageNet-100 \cite{TianKI20} and the challenging ImageNet-1k \cite{RussakovskyDSKS15}.}
    
        \vspace{-3mm}
    \resizebox{\linewidth}{!}{
    \begin{tabular}{ccccccc}
    \toprule
       &\multicolumn{3}{c}{ImageNet-100}&\multicolumn{3}{c}{ImageNet-1k} \\ \cmidrule(lr){2-4} \cmidrule(lr){5-7}
        Method &All&Old&New &All&Old&New \\ \hline
        \textit{k}-means\cite{macqueen1967some} & 72.7 & 75.5 & 71.3 & - &-  &- \\
        RS+\cite{HanREVZ22} & 37.1 & 61.6 & 24.8 & - &-  &-  \\
        UNO+\cite{FiniSLZN021} & 70.3 & 95.0 & 57.9 & - &-  &-  \\
        ORCA\cite{LiuWZFYS23} & 73.5 & 92.6 & 63.9 & - &-  &-  \\ \hline
        GCD  \cite{VazeHVZ22}& 74.1 & 89.8 & 66.3 & 52.5 & 72.5 & 42.2\\
        SimGCD \cite{wen2023parametric} & 83.0 & 93.1 & 77.9 & 57.1 & 77.3 & 46.9\\
        \rowcolor{resultbgsim}
        SimGCD$^{*}$ \cite{wen2023parametric} & 83.3 &92.4 &78.6 &62.3 &79.1 &53.8 \\ \hline
        \rowcolor{resultbg}
        Ours & \textbf{86.3} &\textbf{94.5} &\textbf{82.1}  & \textbf{62.4}	 &\textbf{79.5} &\textbf{53.8}\\
        \rowcolor{resultbg}
        $\Delta$ &\textbf{3.0} &{\color{red}\textbf{2.1}} &\textbf{3.5} &\textbf{0.1} &{\color{red}\textbf{0.4}}&\textbf{0.0}  \\
    \bottomrule
    \end{tabular}}
    \label{tab:result_legegcd_ImageNet}
\end{table}

\textbf{Implementation details.}
Following \cite{VazeHVZ22, wen2023parametric}, we conduct our experiments using ViT-B/16 backbone \cite{DosovitskiyB0WZ21}, which was pre-trained with DINO \cite{CaronTMJMBJ21}, and only fine-tune the last attention block of the backbone for all models. We use the [CLS] token output as the image feature and input for classifier training in SimGCD. Our training regimen includes an initial learning rate of 0.1, decay using a cosine schedule. For a fair comparison, we use a batch size of 128 and train the models for 200 epochs, setting $\lambda=0.35$ in the loss function~(see \cref{loss_cls}). Temperature values $\tau_u=0.07$ and $\tau_c=1.0$ are employed in representation learning. As for training the classifier in SimGCD, we follow the same settings, including $\tau_s=0.1$ and initial $\tau_t=0.07$ which is warmed up to 0.04 with a cosine schedule within the first 30 epochs. In LegoGCD, we set $\tau_{o}=0.05$~(see \cref{loss_ler_entropy}). All experiments are conducted using PyTorch and trained on Nvidia Tesla V100 GPUs.


\subsection{Comparison with the baselines}
We compare our approach with Generalized Category Discovery methods like \textit{k}-means \cite{macqueen1967some}, ORCA\cite{LiuWZFYS23}, GCD \cite{VazeHVZ22}, SimGCD \cite{wen2023parametric}, and strong baselines derived from Novel Category Discovery, including RS+ \cite{HanREVZ22} and UNO+\cite{FiniSLZN021}. For a fair comparison, we reproduce SimGCD~(denoted as SimGCD$^{*}$)~using the same random seed~(\ie seed=0)~as our method. \cref{tab:result_legegcd_1} shows results on SSB datasets \cite{Vaze0VZ22} and Herbarium 19 \cite{tan2019herbarium},  \cref{tab:result_legegcd_CIFAR} and \cref{tab:result_legegcd_ImageNet} present results on generic recognition datasets, including the challenging ImageNet-1k \cite{RussakovskyDSKS15}.

Overall, our method effectively mitigates catastrophic forgetting and achieves superior performance compared to the GCD and SimGCD baseline, particularly in recognizing ``Old" categories. Specifically, it outperforms the baseline by \textbf{3.5\%}/\textbf{2.1\%} on CIFAR100 and ImageNet-100 and shows significant improvements in fine-grained evaluations with \textbf{5.4\%} in CUB and \textbf{5.1\%} in Stanford Cars. Additionally, it surpasses the baseline by \textbf{0.4\%}/\textbf{0.5\%} on the challenging datasets ImageNet-1k and Herbarium 19. In addressing the forgetting problem, our approach also competes well with SimGCD on ``New" classes, achieving \textbf{3.4\%} in Stanford Cars, \textbf{1.0\%} in CIFAR100, and \textbf{3.5\%} in ImageNet-100.

\begin{figure*}[t]
	\centering
	\includegraphics[width=1\linewidth]{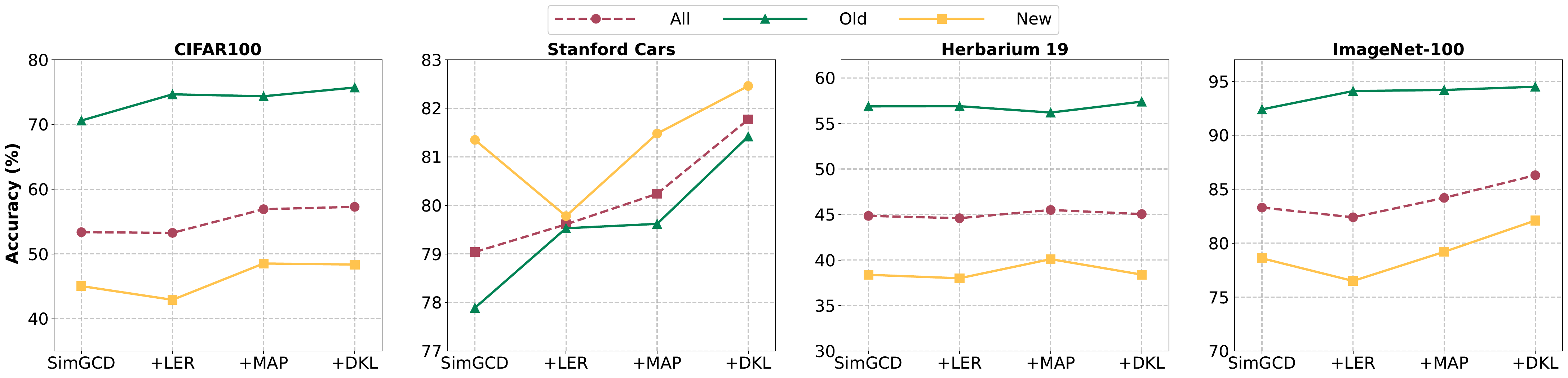}
	\caption{Step by step, we integrate LER and DKL into the baseline SimGCD \cite{wen2023parametric}. Initially, the addition of LER increases accuracy in the ``Old" category while decreasing accuracy in the ``New" category. Subsequently, the introduction of a Margin-aware Pattern (MAP) widens margins between novel categories, ultimately achieving the best performance when embedding with DKL.}
	\label{fig:abs_comps}
\end{figure*}


\subsection{Ablation Study}
In this section, we conduct ablation studies to validate the effectiveness of LER and DKL in LegoGCD. The datasets considered include fine-grained datasets like CUB and Stanford Cars, as well as generic image recognition datasets CIFAR100 and ImageNet-100.
\begin{table}[t]
    \caption{Ablation study on different combinations of our algorithms on CUB. \textbf{Bold} indicates the best results.}
        \vspace{-3mm}
    \resizebox{\linewidth}{!}{
    \begin{tabular}{ccccccc}
    \toprule
         &  &  &  &\multicolumn{3}{c}{CUB} \\  \cmidrule(lr){5-7} 
       \multirow{-2}{*}{SimGCD} &\multirow{-2}{*}{LER}  &\multirow{-2}{*}{MAP} &\multirow{-2}{*}{DKL}  &All  &Old $\uparrow$   &New \\ \hline
        \ding{51} &  &  &  &61.9 &66.5 &59.6 \\
        \ding{51} &  &  & \ding{51} &62.5	&67.3	& \textbf{60.2} \\
        \ding{51} & \ding{51} &  & &63.4 	&71.0	&59.6\\
        \ding{51} & \ding{51} & \ding{51} &  &63.7 &\textbf{72.0} &58.5 \\
        \ding{51} & \ding{51} & \ding{51} & \ding{51} &\textbf{63.8} &{71.9} &{59.8} \\
    \bottomrule
    \end{tabular}}
    \vspace{-2mm}
    \label{tab:abs_CUB}
\end{table}

\textbf{Local Entropy Regularization~(LER).} We conduct ablation studies on LER as shown in \cref{fig:abs_comps}. Firstly, the SimGCD baseline with LER~(without Margin-aware Pattern, MAP)~notably enhances ``Old" Categories (see {\color{graphgreen}green} curves). However, using raw LER alone may impact accuracy in ``New" classes~(see {\color{YellowOrange}orange} curves). This is because it primarily encourages the network to remember known classes, consequently influencing the learning of novel ones.  To mitigate this, we incorporate MAP~(see \cref{MAP}) to encourage the network to simultaneously enhance the margins of all classes, particularly novel classes. Finally, the inclusion of MAP in LER leads to improvements in both the ``Old" and ``New" categories.

\textbf{Dual-views Kullback-Leibler divergence~(DKL).} DKL is designed to improve the quality of known samples for LER by aligning the predictions of two views from the same image. The results in \cref{tab:abs_CUB} indicate that incorporating DKL into raw SimGCD led to improvements in both ``Old" and ``New" category accuracies by 1.8\% and 0.6\%, respectively~(67.3\% vs. 66.5\%, 60.2\% vs. 59.6\%). This success proves beneficial for self-distillation learning in SimGCD. Additionally, from \cref{fig:abs_comps} and \cref{tab:abs_CUB}, we can see that our method becomes more effective with DKL, particularly in ``Old" classes, while maintaining performance on ``New" classes.

\begin{table}[t]
    \caption{Ablation study on $\alpha$ and $\beta$ was conducted on CUB and CIFAR100. \textbf{Bold} indicates the best results. The \underline{underline} denotes the selected $\beta$.}
        \vspace{-3mm}
    \resizebox{\linewidth}{!}{
    \begin{tabular}{ccccccccc}
    \toprule
         & &\multicolumn{3}{c}{CUB} &\multicolumn{3}{c}{CIFAR100} \\ \cmidrule(lr){3-5} \cmidrule(lr){6-8}
        $\alpha$ &{$\beta$} &All  &Old $\uparrow$  &New   &All  &Old $\uparrow$   &New \\ \hline
         & 0.0 &61.9 &66.5 &59.6 &79.0 &77.9 &81.5 \\
         & 0.5 &63.0 &69.5 &59.7   &80.8 &80.4 &81.5\\
         & \underline{1.0} &63.6 &69.7	&\textbf{60.6}  &\textbf{81.8} & 81.4 & \textbf{82.5} \\
         & 1.5 &63.4 &71.2 &59.8   &81.4 &81.7  &80.8 \\
        \multirow{-3}{*}{1.0} &\underline{2.0} & \textbf{63.8} & \textbf{71.9} & 59.8 &  81.7  &\textbf{82.1}  & 80.9 \\
    \bottomrule
    \end{tabular}}
        \vspace{-3mm}
    \label{tab:abs_different_hyp}
\end{table}
\textbf{Different $\alpha$ and $\beta$.}
The coefficient $\beta$ is crucial for balancing knowledge preservation of known classes and facilitating effective learning of novel classes. As depicted in \cref{tab:abs_different_hyp} with $\alpha=1.0$ aligned to SimGCD, accuracy on ``Old" categories consistently surpasses SimGCD ($\beta=0.0$) across different $\beta$ values. Optimal equilibrium is achieved with $\beta=2.0$ in CUB and $\beta=1.0$ in CIFAR100.

\textbf{Different confidence threshold $\delta$.}
We compare accuracy with different $\delta$ in \cref{tab:abs_different_thr}. Results show varying responses across datasets. In Stanford Cars, ``Old" accuracy initially increases, then slightly decreases, while ``New" accuracy decreases but remains above SimGCD at 45.5\%. In contrast, in CIFAR100, ``Old" accuracy decreases but consistently surpasses SimGCD at 77.9\%. Our goal is to prioritize high ``Old" accuracy and ensure ``New" equals or exceeds SimGCD. Therefore, we choose $\delta=0.85$ for both Stanford Cars and CIFAR100.
\begin{table}[t]
    \caption{Ablation study on confidence threshold $\delta$ was conducted on Stanford Cars and CIFAR100. \textbf{Bold} indicates the best results. The \underline{underline} denotes the selected threshold.}
    \vspace{-3mm}
    \small
    \centering
    \resizebox{\linewidth}{!}{
    \begin{tabular}{ccccccc}
    \toprule
        & \multicolumn{3}{c}{Stanford Cars} &\multicolumn{3}{c}{CIFAR100}\\ \cmidrule(lr){2-4} \cmidrule(lr){5-7}
        $\delta$ &All  &Old $\uparrow$ &New   &All  &Old $\downarrow$ &New  \\ \hline
        0.70 &55.3	&66.8  &\textbf{49.8} & 80.3  &\textbf{83.3} &74.3 \\
        0.75 &55.8	&72.3  &47.9 &80.9	&83.3	&76.3 \\
        0.80 &56.5	&73.0  &48.5 &81.3	&82.7	&78.5 \\
        \underline{0.85} &\textbf{57.3} &\textbf{75.7} &48.4 &\textbf{81.8} 	&81.4	&\textbf{82.5} \\
        0.90 &56.4  &75.3  &47.3  &79.0	&80.5	&76.0 \\
    \bottomrule
    \end{tabular}}
        \vspace{-3mm}
    \label{tab:abs_different_thr}
\end{table}

\section{Conclusion}
In this paper, we propose LegoGCD, a novel approach to mitigate the issue of catastrophic forgetting in known categories. The core of our design is to preserve knowledge of known classes while maintaining the accuracy of novel classes. To achieve this, we develop two techniques: Local Entropy Regularization~(LER)~and Dual-views Kullback-Leibler divergence constraint~(DKL). LER explicitly regularizes high-confidence potential known class samples to retain the knowledge of known categories. To ensure the accurate selection of these samples, we employ DKL to align the distribution of two view samples for LER. Both LER and DKL can be seamlessly integrated into baseline SimGCD resembling Lego blocks, without introducing new parameters or altering the internal network structure. Extensive experiments demonstrate that LegoGCD significantly enhances performance in known classes, effectively addressing the catastrophic forgetting problem.

\section{Acknowledgements}
This research is supported by the National Key R\&D Program of China (2021YFB0301300), and also received supported from programs: the Major Program of Guangdong Basic and Applied Research (2019B030302002), the Key-Area Research and Development Program of Guangdong Province (2021B0101400002), the Major Key Project of PCL PCL2021A13 and Peng Cheng Cloud-Brain, and the Fundamental Research Funds for the Central Universities, Sun Yat-sen University (23xkjc016).

{
    \small
    \bibliographystyle{ieeenat_fullname}
    \bibliography{main}

\begin{thebibliography}{51}
\providecommand{\natexlab}[1]{#1}
\providecommand{\url}[1]{\texttt{#1}}
\expandafter\ifx\csname urlstyle\endcsname\relax
  \providecommand{\doi}[1]{doi: #1}\else
  \providecommand{\doi}{doi: \begingroup \urlstyle{rm}\Url}\fi

\bibitem[An et~al.(2023)An, Tian, Zheng, Ding, Wang, and Chen]{an2023generalized}
Wenbin An, Feng Tian, Qinghua Zheng, Wei Ding, QianYing Wang, and Ping Chen.
\newblock Generalized category discovery with decoupled prototypical network.
\newblock In \emph{Proceedings of the AAAI Conference on Artificial Intelligence (AAAI)}, pages 12527--12535, 2023.

\bibitem[Assran et~al.(2022)Assran, Caron, Misra, Bojanowski, Bordes, Vincent, Joulin, Rabbat, and Ballas]{AssranCMBBVJRB22}
Mahmoud Assran, Mathilde Caron, Ishan Misra, Piotr Bojanowski, Florian Bordes, Pascal Vincent, Armand Joulin, Mike Rabbat, and Nicolas Ballas.
\newblock Masked siamese networks for label-efficient learning.
\newblock In \emph{Proceedings of the European conference on Computer Vision (ECCV)}, pages 456--473, 2022.

\bibitem[Berthelot et~al.(2019)Berthelot, Carlini, Goodfellow, Papernot, Oliver, and Raffel]{BerthelotCGPOR19}
David Berthelot, Nicholas Carlini, Ian~J. Goodfellow, Nicolas Papernot, Avital Oliver, and Colin Raffel.
\newblock Mixmatch: {A} holistic approach to semi-supervised learning.
\newblock In \emph{Advances in Neural Information Processing Systems (NeurIPS)}, pages 5050--5060, 2019.

\bibitem[Caron et~al.(2021)Caron, Touvron, Misra, J{\'{e}}gou, Mairal, Bojanowski, and Joulin]{CaronTMJMBJ21}
Mathilde Caron, Hugo Touvron, Ishan Misra, Herv{\'{e}} J{\'{e}}gou, Julien Mairal, Piotr Bojanowski, and Armand Joulin.
\newblock Emerging properties in self-supervised vision transformers.
\newblock In \emph{Proceedings of {IEEE/CVF} International Conference on Computer Vision (CVPR)}, pages 9630--9640, 2021.

\bibitem[Deng et~al.(2009)Deng, Dong, Socher, Li, Li, and Fei{-}Fei]{DengDSLL009}
Jia Deng, Wei Dong, Richard Socher, Li{-}Jia Li, Kai Li, and Li Fei{-}Fei.
\newblock Imagenet: {A} large-scale hierarchical image database.
\newblock In \emph{Proceedings of {IEEE/CVF} Conference on Computer Vision and Pattern Recognition (CVPR)}, pages 248--255, 2009.

\bibitem[Dosovitskiy et~al.(2021)Dosovitskiy, Beyer, Kolesnikov, Weissenborn, Zhai, Unterthiner, Dehghani, Minderer, Heigold, Gelly, Uszkoreit, and Houlsby]{DosovitskiyB0WZ21}
Alexey Dosovitskiy, Lucas Beyer, Alexander Kolesnikov, Dirk Weissenborn, Xiaohua Zhai, Thomas Unterthiner, Mostafa Dehghani, Matthias Minderer, Georg Heigold, Sylvain Gelly, Jakob Uszkoreit, and Neil Houlsby.
\newblock An image is worth 16x16 words: Transformers for image recognition at scale.
\newblock In \emph{Proceedings of International Conference on Learning Representations (ICLR)}, 2021.

\bibitem[Fini et~al.(2021)Fini, Sangineto, Lathuili{\`{e}}re, Zhong, Nabi, and Ricci]{FiniSLZN021}
Enrico Fini, Enver Sangineto, St{\'{e}}phane Lathuili{\`{e}}re, Zhun Zhong, Moin Nabi, and Elisa Ricci.
\newblock A unified objective for novel class discovery.
\newblock In \emph{Proceedings of {IEEE/CVF} International Conference on Computer Vision (ICCV)}, pages 9264--9272, 2021.

\bibitem[Han et~al.(2019)Han, Vedaldi, and Zisserman]{HanVZ19}
Kai Han, Andrea Vedaldi, and Andrew Zisserman.
\newblock Learning to discover novel visual categories via deep transfer clustering.
\newblock In \emph{Proceedings of {IEEE/CVF} International Conference on Computer Vision (ICCV)}, pages 8400--8408, 2019.

\bibitem[Han et~al.(2022)Han, Rebuffi, Ehrhardt, Vedaldi, and Zisserman]{HanREVZ22}
Kai Han, Sylvestre{-}Alvise Rebuffi, S{\'{e}}bastien Ehrhardt, Andrea Vedaldi, and Andrew Zisserman.
\newblock Autonovel: Automatically discovering and learning novel visual categories.
\newblock \emph{{IEEE} Trans. Pattern Anal. Mach. Intell.}, 44\penalty0 (10):\penalty0 6767--6781, 2022.

\bibitem[He et~al.(2016)He, Zhang, Ren, and Sun]{HeZRS16}
Kaiming He, Xiangyu Zhang, Shaoqing Ren, and Jian Sun.
\newblock Deep residual learning for image recognition.
\newblock In \emph{Proceedings of the {IEEE/CVF} Conference on Computer Vision and Pattern Recognition (CVPR)}, pages 770--778, 2016.

\bibitem[He et~al.(2017)He, Gkioxari, Doll{\'{a}}r, and Girshick]{HeGDG17}
Kaiming He, Georgia Gkioxari, Piotr Doll{\'{a}}r, and Ross~B. Girshick.
\newblock Mask {R-CNN}.
\newblock In \emph{Proceedings of the {IEEE/CVF} Conference on Computer Vision and Pattern Recognition (CVPR)}, pages 2980--2988, 2017.

\bibitem[Huang et~al.(2017)Huang, Liu, van~der Maaten, and Weinberger]{HuangLMW17}
Gao Huang, Zhuang Liu, Laurens van~der Maaten, and Kilian~Q. Weinberger.
\newblock Densely connected convolutional networks.
\newblock In \emph{Proceedings of the {IEEE/CVF} Conference on Computer Vision and Pattern Recognition (CVPR)}, pages 2261--2269, 2017.

\bibitem[Joseph et~al.(2022)Joseph, Paul, Aggarwal, Biswas, Rai, Han, and Balasubramanian]{JosephPABRHB22}
K.~J. Joseph, Sujoy Paul, Gaurav Aggarwal, Soma Biswas, Piyush Rai, Kai Han, and Vineeth~N. Balasubramanian.
\newblock Novel class discovery without forgetting.
\newblock In \emph{Proceedings of the European conference on Computer Vision (ECCV)}, pages 570--586, 2022.

\bibitem[Krause et~al.(2013)Krause, Stark, Deng, and Fei{-}Fei]{Krause0DF13}
Jonathan Krause, Michael Stark, Jia Deng, and Li Fei{-}Fei.
\newblock 3d object representations for fine-grained categorization.
\newblock In \emph{Proceedings of {IEEE/CVF} International Conference on Computer Vision Workshops (ICCV)}, pages 554--561, 2013.

\bibitem[Krizhevsky et~al.(2009)Krizhevsky, Hinton, et~al.]{krizhevsky2009learning}
Alex Krizhevsky, Geoffrey Hinton, et~al.
\newblock Learning multiple layers of features from tiny images.
\newblock 2009.

\bibitem[Kuhn(1955)]{kuhn1955hungarian}
Harold~W Kuhn.
\newblock The hungarian method for the assignment problem.
\newblock \emph{Naval research logistics quarterly}, 2\penalty0 (1-2):\penalty0 83--97, 1955.

\bibitem[Laine and Aila(2017)]{LaineA17}
Samuli Laine and Timo Aila.
\newblock Temporal ensembling for semi-supervised learning.
\newblock In \emph{Proceedings of International Conference on Learning Representations (ICLR)}. OpenReview.net, 2017.

\bibitem[Li et~al.(2023)Li, Fan, Huo, and Gao]{LiFHG23}
Wenbin Li, Zhichen Fan, Jing Huo, and Yang Gao.
\newblock Modeling inter-class and intra-class constraints in novel class discovery.
\newblock In \emph{Proceedings of {IEEE/CVF} Conference on Computer Vision and Pattern Recognition (CVPR)}, pages 3449--3458, 2023.

\bibitem[Liu et~al.(2023)Liu, Wang, Zhang, Fan, Yang, and Shao]{LiuWZFYS23}
Jiaming Liu, Yangqiming Wang, Tongze Zhang, Yulu Fan, Qinli Yang, and Junming Shao.
\newblock Open-world semi-supervised novel class discovery.
\newblock In \emph{Proceedings of the Thirty-Second International Joint Conference on Artificial Intelligence, {IJCAI}}, pages 4002--4010. ijcai.org, 2023.

\bibitem[MacQueen et~al.(1967)]{macqueen1967some}
James MacQueen et~al.
\newblock Some methods for classification and analysis of multivariate observations.
\newblock In \emph{Proceedings of the fifth Berkeley symposium on mathematical statistics and probability}, pages 281--297, 1967.

\bibitem[Maji et~al.(2013)Maji, Rahtu, Kannala, Blaschko, and Vedaldi]{MajiRKBV13}
Subhransu Maji, Esa Rahtu, Juho Kannala, Matthew~B. Blaschko, and Andrea Vedaldi.
\newblock Fine-grained visual classification of aircraft.
\newblock \emph{arXiv preprint arXiv:1306.5151}, 2013.

\bibitem[Miyato et~al.(2019)Miyato, Maeda, Koyama, and Ishii]{MiyatoMKI19}
Takeru Miyato, Shin{-}ichi Maeda, Masanori Koyama, and Shin Ishii.
\newblock Virtual adversarial training: {A} regularization method for supervised and semi-supervised learning.
\newblock \emph{{IEEE} Trans. Pattern Anal. Mach. Intell.}, 41\penalty0 (8):\penalty0 1979--1993, 2019.

\bibitem[Pu et~al.(2023)Pu, Zhong, and Sebe]{PuZS23}
Nan Pu, Zhun Zhong, and Nicu Sebe.
\newblock Dynamic conceptional contrastive learning for generalized category discovery.
\newblock In \emph{Proceedings of the {IEEE/CVF} Conference on Computer Vision and Pattern Recognition (CVPR)}, pages 7579--7588, 2023.

\bibitem[Ren et~al.(2015)Ren, He, Girshick, and Sun]{RenHGS15}
Shaoqing Ren, Kaiming He, Ross~B. Girshick, and Jian Sun.
\newblock Faster {R-CNN:} towards real-time object detection with region proposal networks.
\newblock In \emph{Advances in Neural Information Processing Systems (NeurIPS)}, pages 91--99, 2015.

\bibitem[Ronneberger et~al.(2015)Ronneberger, Fischer, and Brox]{DRonnebergerFB15}
Olaf Ronneberger, Philipp Fischer, and Thomas Brox.
\newblock U-net: Convolutional networks for biomedical image segmentation.
\newblock In \emph{Medical Image Computing and Computer-Assisted Intervention (MICCAI)}, pages 234--241, 2015.

\bibitem[Russakovsky et~al.(2015)Russakovsky, Deng, Su, Krause, Satheesh, Ma, Huang, Karpathy, Khosla, Bernstein, Berg, and Fei{-}Fei]{RussakovskyDSKS15}
Olga Russakovsky, Jia Deng, Hao Su, Jonathan Krause, Sanjeev Satheesh, Sean Ma, Zhiheng Huang, Andrej Karpathy, Aditya Khosla, Michael~S. Bernstein, Alexander~C. Berg, and Li Fei{-}Fei.
\newblock Imagenet large scale visual recognition challenge.
\newblock \emph{Int. J. Comput. Vis.}, 115\penalty0 (3):\penalty0 211--252, 2015.

\bibitem[Sandler et~al.(2018)Sandler, Howard, Zhu, Zhmoginov, and Chen]{SandlerHZZC18}
Mark Sandler, Andrew~G. Howard, Menglong Zhu, Andrey Zhmoginov, and Liang{-}Chieh Chen.
\newblock Mobilenetv2: Inverted residuals and linear bottlenecks.
\newblock In \emph{Proceedings of {IEEE/CVF} Conference on Computer Vision and Pattern Recognition (CVPR)}, pages 4510--4520, 2018.

\bibitem[Simonyan and Zisserman(2014)]{simonyan2014very}
Karen Simonyan and Andrew Zisserman.
\newblock Very deep convolutional networks for large-scale image recognition.
\newblock \emph{arXiv preprint arXiv:1409.1556}, 2014.

\bibitem[Sohn et~al.(2020)Sohn, Berthelot, Carlini, Zhang, Zhang, Raffel, Cubuk, Kurakin, and Li]{SohnBCZZRCKL20}
Kihyuk Sohn, David Berthelot, Nicholas Carlini, Zizhao Zhang, Han Zhang, Colin Raffel, Ekin~Dogus Cubuk, Alexey Kurakin, and Chun{-}Liang Li.
\newblock Fixmatch: Simplifying semi-supervised learning with consistency and confidence.
\newblock In \emph{Advances in Neural Information Processing Systems (NeurIPS)}, 2020.

\bibitem[Szegedy et~al.(2015)Szegedy, Liu, Jia, Sermanet, Reed, Anguelov, Erhan, Vanhoucke, and Rabinovich]{SzegedyLJSRAEVR15}
Christian Szegedy, Wei Liu, Yangqing Jia, Pierre Sermanet, Scott~E. Reed, Dragomir Anguelov, Dumitru Erhan, Vincent Vanhoucke, and Andrew Rabinovich.
\newblock Going deeper with convolutions.
\newblock In \emph{Proceedings of the {IEEE/CVF} Conference on Computer Vision and Pattern Recognition (CVPR)}, pages 1--9, 2015.

\bibitem[Tan et~al.(2019)Tan, Liu, Ambrose, Tulig, and Belongie]{tan2019herbarium}
Kiat~Chuan Tan, Yulong Liu, Barbara Ambrose, Melissa Tulig, and Serge Belongie.
\newblock The herbarium challenge 2019 dataset.
\newblock \emph{arXiv preprint arXiv:1906.05372}, 2019.

\bibitem[Tarvainen and Valpola(2017)]{TarvainenV17}
Antti Tarvainen and Harri Valpola.
\newblock Mean teachers are better role models: Weight-averaged consistency targets improve semi-supervised deep learning results.
\newblock In \emph{Advances in Neural Information Processing Systems (NeurIPS)}, pages 1195--1204, 2017.

\bibitem[Tian et~al.(2020)Tian, Krishnan, and Isola]{TianKI20}
Yonglong Tian, Dilip Krishnan, and Phillip Isola.
\newblock Contrastive multiview coding.
\newblock In \emph{Proceedings of the European conference on Computer Vision (ECCV)}, pages 776--794, 2020.

\bibitem[Vaswani et~al.(2017)Vaswani, Shazeer, Parmar, Uszkoreit, Jones, Gomez, Kaiser, and Polosukhin]{VaswaniSPUJGKP17}
Ashish Vaswani, Noam Shazeer, Niki Parmar, Jakob Uszkoreit, Llion Jones, Aidan~N. Gomez, Lukasz Kaiser, and Illia Polosukhin.
\newblock Attention is all you need.
\newblock In \emph{Proceedings of Advances in Neural Information Processing Systems (NeurIPS)}, pages 5998--6008, 2017.

\bibitem[Vaze et~al.(2022{\natexlab{a}})Vaze, Han, Vedaldi, and Zisserman]{Vaze0VZ22}
Sagar Vaze, Kai Han, Andrea Vedaldi, and Andrew Zisserman.
\newblock Open-set recognition: {A} good closed-set classifier is all you need.
\newblock In \emph{Proceedings of International Conference on Learning Representations (ICLR)}, 2022{\natexlab{a}}.

\bibitem[Vaze et~al.(2022{\natexlab{b}})Vaze, Han, Vedaldi, and Zisserman]{VazeHVZ22}
Sagar Vaze, Kai Han, Andrea Vedaldi, and Andrew Zisserman.
\newblock Generalized category discovery.
\newblock In \emph{Proceedings of the {IEEE/CVF} Conference on Computer Vision and Pattern Recognition (CVPR)}, pages 7482--7491, 2022{\natexlab{b}}.

\bibitem[Wah et~al.(2011)Wah, Branson, Welinder, Perona, and Belongie]{wah2011caltech}
Catherine Wah, Steve Branson, Peter Welinder, Pietro Perona, and Serge Belongie.
\newblock The caltech-ucsd birds-200-2011 dataset.
\newblock 2011.

\bibitem[Wang et~al.(2022)Wang, Wu, Lian, and Yu]{WangWLY22}
Xudong Wang, Zhirong Wu, Long Lian, and Stella~X. Yu.
\newblock Debiased learning from naturally imbalanced pseudo-labels.
\newblock In \emph{Proceedings of {IEEE/CVF} Conference on Computer Vision and Pattern Recognition (CVPR)}, pages 14627--14637, 2022.

\bibitem[Wen et~al.(2023)Wen, Zhao, and Qi]{wen2023parametric}
Xin Wen, Bingchen Zhao, and Xiaojuan Qi.
\newblock Parametric classification for generalized category discovery: A baseline study.
\newblock In \emph{Proceedings of the IEEE/CVF International Conference on Computer Vision (ICCV)}, pages 16590--16600, 2023.

\bibitem[Yang et~al.(2022)Yang, Zhu, Yu, Wu, and Deng]{YangZYWD22}
Muli Yang, Yuehua Zhu, Jiaping Yu, Aming Wu, and Cheng Deng.
\newblock Divide and conquer: Compositional experts for generalized novel class discovery.
\newblock In \emph{Proceedings of {IEEE/CVF} Conference on Computer Vision and Pattern Recognition (CVPR)}, pages 14248--14257, 2022.

\bibitem[Yang et~al.(2023)Yang, Wang, Deng, and Zhang]{YangWDZ23}
Muli Yang, Liancheng Wang, Cheng Deng, and Hanwang Zhang.
\newblock Bootstrap your own prior: Towards distribution-agnostic novel class discovery.
\newblock In \emph{Proceedings of {IEEE/CVF} Conference on Computer Vision and Pattern Recognition (CVPR)}, pages 3459--3468, 2023.

\bibitem[Yu et~al.(2023)Yu, Li, and Lee]{YuLL23}
Zhuoran Yu, Yin Li, and Yong~Jae Lee.
\newblock Inpl: Pseudo-labeling the inliers first for imbalanced semi-supervised learning.
\newblock In \emph{Proceedings of International Conference on Learning Representations (ICLR)}, 2023.

\bibitem[Yun et~al.(2019)Yun, Han, Chun, Oh, Yoo, and Choe]{YunHCOYC19}
Sangdoo Yun, Dongyoon Han, Sanghyuk Chun, Seong~Joon Oh, Youngjoon Yoo, and Junsuk Choe.
\newblock Cutmix: Regularization strategy to train strong classifiers with localizable features.
\newblock In \emph{Proceedings of {IEEE/CVF} International Conference on Computer Vision (ICCV)}, pages 6022--6031, 2019.

\bibitem[Zhang et~al.(2018)Zhang, Ciss{\'{e}}, Dauphin, and Lopez{-}Paz]{ZhangCDL18}
Hongyi Zhang, Moustapha Ciss{\'{e}}, Yann~N. Dauphin, and David Lopez{-}Paz.
\newblock mixup: Beyond empirical risk minimization.
\newblock In \emph{Proceedings of International Conference on Learning Representations (ICLR)}, 2018.

\bibitem[Zhang et~al.(2023{\natexlab{a}})Zhang, Ma, Guo, and Xu]{ZhangMG023}
Jie Zhang, Xiaosong Ma, Song Guo, and Wenchao Xu.
\newblock Towards unbiased training in federated open-world semi-supervised learning.
\newblock In \emph{Proceedings of International Conference on Machine Learning, (ICML)}, pages 41498--41509, 2023{\natexlab{a}}.

\bibitem[Zhang et~al.(2023{\natexlab{b}})Zhang, Khan, Shen, Naseer, Chen, and Khan]{ZhangKSNCK23}
Sheng Zhang, Salman~H. Khan, Zhiqiang Shen, Muzammal Naseer, Guangyi Chen, and Fahad~Shahbaz Khan.
\newblock Promptcal: Contrastive affinity learning via auxiliary prompts for generalized novel category discovery.
\newblock In \emph{Proceedings of the {IEEE/CVF} Conference on Computer Vision and Pattern Recognition (CVPR)}, pages 3479--3488, 2023{\natexlab{b}}.

\bibitem[Zhang et~al.(2022)Zhang, Zhu, Hallinan, Zhang, Makmur, Cai, and Ooi]{ZhangZHZMCO22}
Wenqiao Zhang, Lei Zhu, James Hallinan, Shengyu Zhang, Andrew Makmur, Qingpeng Cai, and Beng~Chin Ooi.
\newblock Boostmis: Boosting medical image semi-supervised learning with adaptive pseudo labeling and informative active annotation.
\newblock In \emph{Proceedings of {IEEE/CVF} Conference on Computer Vision and Pattern Recognition (CVPR)}, pages 20634--20644, 2022.

\bibitem[Zhao and Han(2021)]{ZhaoH21}
Bingchen Zhao and Kai Han.
\newblock Novel visual category discovery with dual ranking statistics and mutual knowledge distillation.
\newblock In \emph{Advances in Neural Information Processing Systems (NeurIPS)}, pages 22982--22994, 2021.

\bibitem[Zhao et~al.(2022)Zhao, Zhong, Sebe, and Lee]{ZhaoZSL22}
Yuyang Zhao, Zhun Zhong, Nicu Sebe, and Gim~Hee Lee.
\newblock Novel class discovery in semantic segmentation.
\newblock In \emph{Proceedings of {IEEE/CVF} Conference on Computer Vision and Pattern Recognition (CVPR)}, pages 4330--4339, 2022.

\bibitem[Zhong et~al.(2021)Zhong, Zhu, Luo, Li, Yang, and Sebe]{ZhongZLL0S21}
Zhun Zhong, Linchao Zhu, Zhiming Luo, Shaozi Li, Yi Yang, and Nicu Sebe.
\newblock Openmix: Reviving known knowledge for discovering novel visual categories in an open world.
\newblock In \emph{Proceedings of {IEEE/CVF} Conference on Computer Vision and Pattern Recognition (CVPR)}, pages 9462--9470, 2021.

\bibitem[Zoph et~al.(2018)Zoph, Vasudevan, Shlens, and Le]{ZophVSL18}
Barret Zoph, Vijay Vasudevan, Jonathon Shlens, and Quoc~V. Le.
\newblock Learning transferable architectures for scalable image recognition.
\newblock In \emph{Proceedings of {IEEE/CVF} Conference on Computer Vision and Pattern Recognition (CVPR)}, pages 8697--8710, 2018.

\end{thebibliography}
}
\clearpage
\setcounter{page}{1}
\setcounter{section}{0}
\section{Data details}
\begin{figure}[htb]
	\includegraphics[width=1\linewidth]{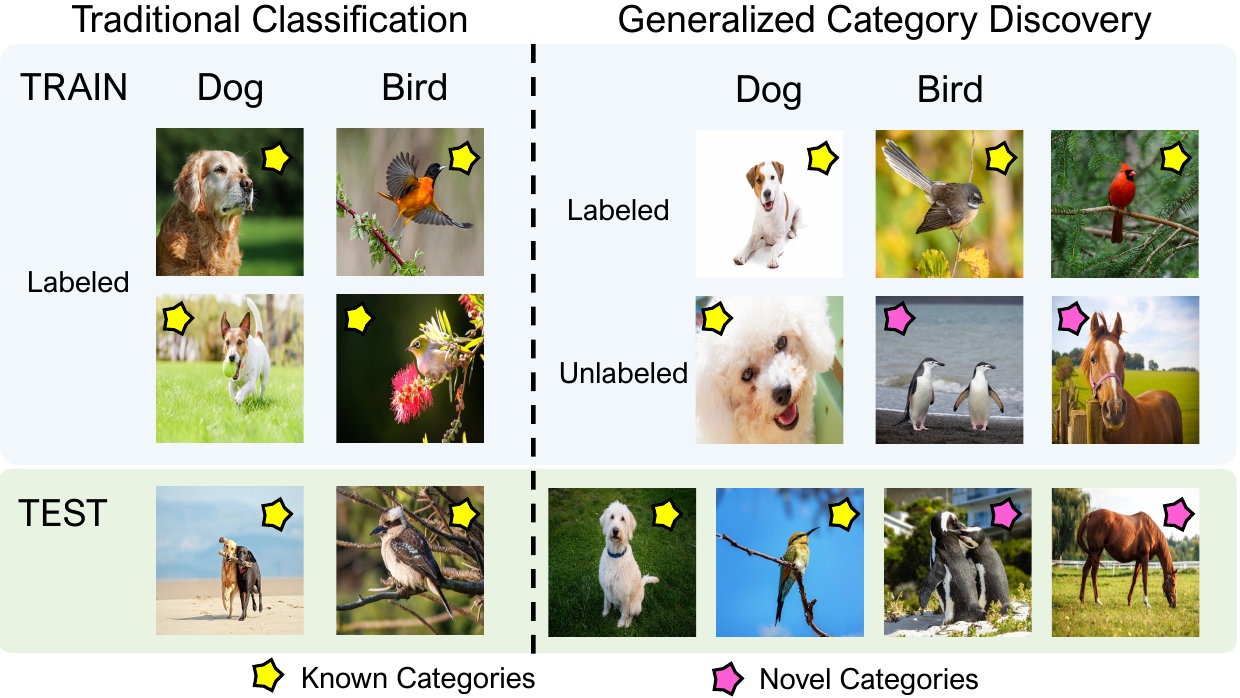}
	\caption{Data details of traditional classification and Generalized Category Discovery.}
	\label{fig:data}
\end{figure}

\cref{fig:data} illustrates the data difference between traditional classification and Generalized Category Discovery (GCD). Unlike traditional classification models trained in a closed set, where both training and test data only come from labeled data, GCD operates in an open set—a more realistic and challenging setting. In GCD, the training data includes unlabeled samples that consist of both known classes (\eg dog and bird) and novel classes (\eg penguin and horse) without annotations. During testing, the model should accurately classify the known class samples and recognize the novel class samples.

\vspace{-5mm}	
\section{Training visualization}
\begin{figure*}[htb]
	\centering
	\includegraphics[width=0.95\linewidth]{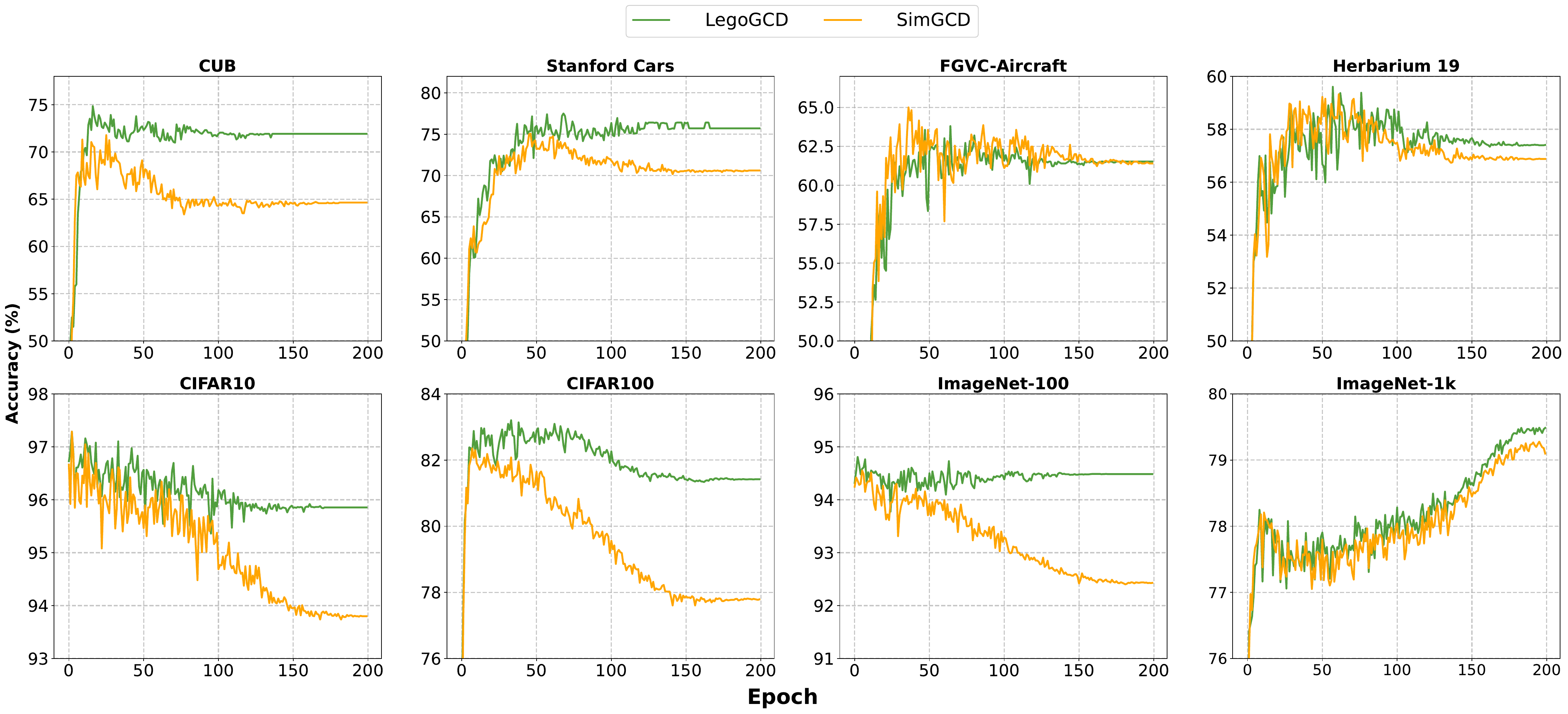}
        \vspace{-3mm}	
 \caption{``Old" accuracy in each epoch compared between SimGCD and our LegoGCD. Our method (depicted in {\color{graphgreen}green}) consistently outperforms SimGCD (shown in {\color{YellowOrange}orange}) across all datasets.}
	\label{fig:training_epochs}
\end{figure*}

\cref{fig:training_epochs} shows the ``Old" accuracy across training epochs for both SimGCD and our LegoGCD, employing the same random seed. Our method (depicted by {\color{graphgreen}green} curves) consistently outperforms SimGCD (shown by {\color{YellowOrange}orange} curves) across diverse datasets. Notably, LegoGCD effectively addresses the catastrophic forgetting problem, particularly in fine-grained datasets like CUB and Stanford Cars, as well as in generic image recognition datasets CIFAR10/100 and ImageNet-100. Meanwhile, LegoGCD enhances known class accuracies, even in datasets with less pronounced forgetting, such as the unbalanced Herbarium 19. Additionally, improvements are observed in FGVA-Aircraft and ImageNet-1k datasets without forgetting.
\begin{figure*}[htb]
   \centering
    \hspace{-4mm}\begin{subfigure}{0.33\linewidth}
   \includegraphics[width=1\linewidth]{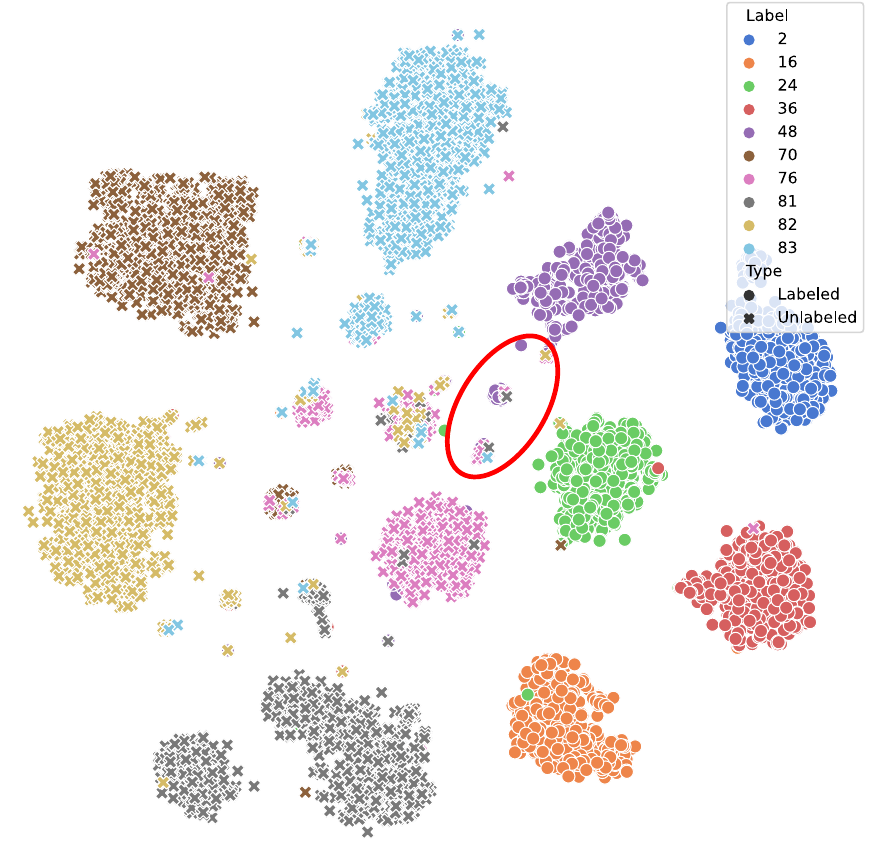}
    \caption{ImageNet-100 on SimGCD}
    \label{fig:ImageNet100_simgcd}
  \end{subfigure}
  \hspace{2mm}\begin{subfigure}{0.33\linewidth}
    \includegraphics[width=1\linewidth]{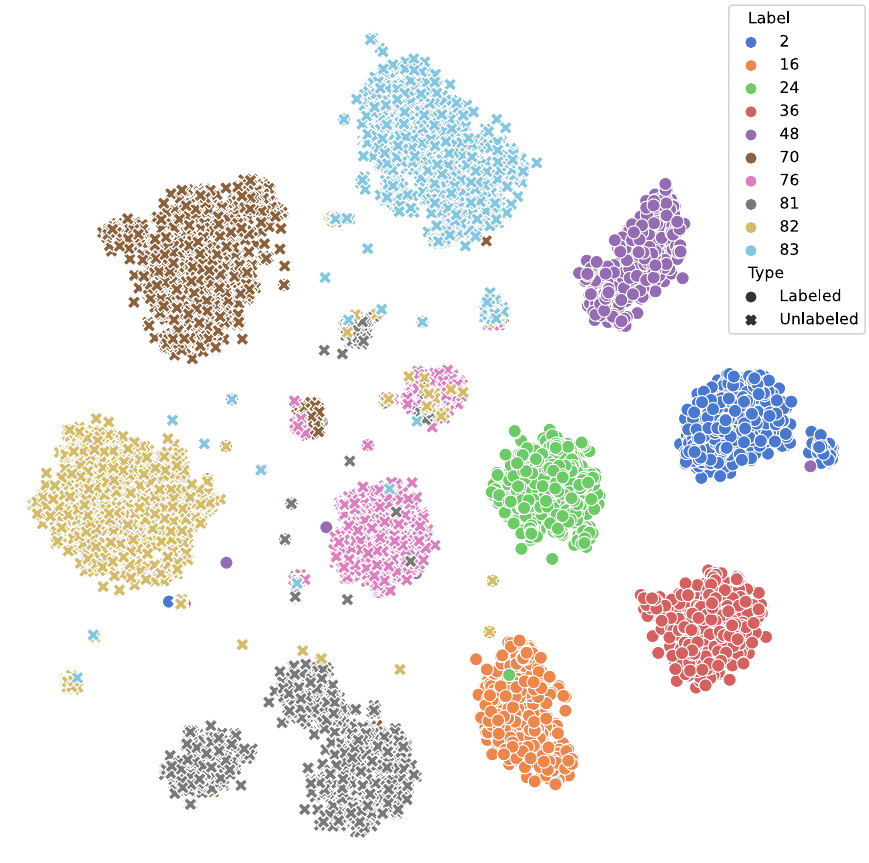}
    \caption{ImageNet-100 on LegoGCD}
    \label{fig:ImageNet100_legogcd}
  \end{subfigure}
  \hspace{3mm}\begin{subfigure}{0.325\linewidth}
    \includegraphics[width=1\linewidth]{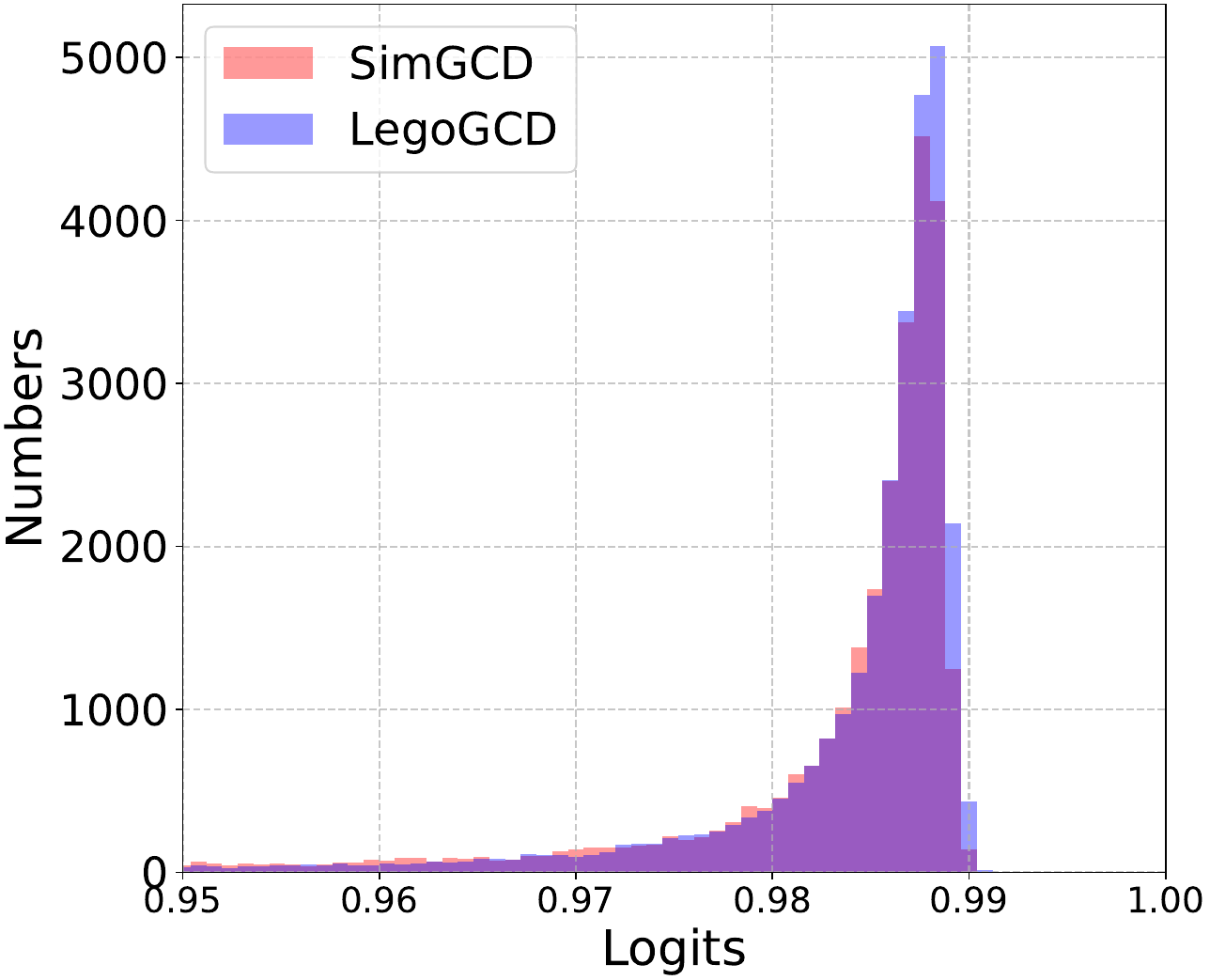}
    \caption{Logits distributions}
    \label{fig:log_dis}
  \end{subfigure}
\caption{The t-SNE visualization and logit distributions of the unlabeled dataset for SimGCD and LegoGCD ImageNet-100.}
	\label{fig:tsne}
\end{figure*}
\begin{algorithm}[t]
        \caption{Pseudo code on one step for LegoGCD} 
        	\label{alg:algorithm} 
        \begin{lstlisting}[language=Python]
#x1, x2: two view samples
#s_proj, s_pred, t_pred: projection feature, logits (similarities) for student and teacher
#mask: label mask
#x1_pred, x2_pred: logits of two view samples
def training_step(x1, x2):
    s_proj, s_pred = model([x1, x2])
    t_pred = s_pred.detach()
    #(1)Representation learning (unsupervised)
    unsup_con_loss = UnsupConLoss(s_proj)#Eq.(1)
    #(2) Representation learning (supervised) 
    sup_con_loss = SupConLoss(s_proj, label=target[mask=1]) #Eq.(2)
    #(3) Supervised classification loss uses ground-truth labels on labeled data
    sup_loss = cross_entropy(s_pred[mask=1], target[mask=1]) #Eq.(5)
    #(4) Unsupervised (Self-distillation) classification loss on all data in Eq.(5)
    unsup_loss = cross_entropy(t_pred, s_pred)
    #(5) DKL
    x1_pred, x2_pred.detach() = s_pred.chunk(2)
    unsup_loss += DKL(x1_pred, x2_pred) #Eq.(11)
    #(6) LER in Eq.(10)
    loss_ler = LER(s_pred, s_pred+delta_logits)
    # Total representation learning loss
    loss_rep =  (1-lambda)*unsup_con_loss + sup_con_loss
    # Total classification loss
    loss_cls = (1-lambda)*unsup_loss + sup_loss
    # Overall loss
    loss = alpha * (loss_rep + loss_cls ) +beta* loss_ler #Eq.(13)
    return loss
\end{lstlisting}
\end{algorithm}
\section{Representations visualization}
In this section, we employ t-distributed stochastic neighbor embedding (t-SNE) to visualize the learned representations of LegoGCD and compare them with the baseline SimGCD. The result of this comparison is presented in \cref{fig:tsne}. Spectively, we randomly select 10 categories, each composed of 5 known and novel classes, with known and novel samples marked with \ding{108} and \ding{54}, respectively. \cref{fig:ImageNet100_simgcd} and \cref{fig:ImageNet100_legogcd} display the visualizations on ImageNet-100 in SimGCD and our method, respectively. In \cref{fig:ImageNet100_simgcd}, some representations of novel classes are closer to known classes 24 and 48 than their truth labels, which are circled by red color. On the contrary, the representations of our method in known categories in  \cref{fig:ImageNet100_legogcd} exhibit clear margins, indicating our method can more effectively distinguish known samples. Furthermore, \cref{fig:log_dis} illustrates the logit distribution of known samples in unlabeled data. The predictions of our method exhibit higher logits, indicating enhanced sample discriminability.

\section{Experimental supplements}
In this section, we give detailed analyses of CIFAR10 and FGVC-Aircraft which improvements are not obvious in ``Old" classes.
\subsection{Results on CIFAR10}
In this section, we conduct an ablation study on the confidence threshold in CIFAR10, as detailed in \cref{tab:cifar10_thr}. Notably, the ``Old" accuracy consistently surpasses that of SimGCD when $\delta\!=\!0$. Despite a marginal drop in ``New" accuracy ranging from 0.3 to 0.5, significant improvements are observed in ``Old" accuracy, effectively mitigating the forgetting problem and show significant robustness in ``Old" classes. Ultimately, we select $\delta=0.97$ as the optimal threshold. While this choice results in a 0.3\% reduction in ``New" accuracy, it boosts ``Old" accuracy by 1.9\%, leading to an overall improvement of 0.6\% in ``All" accuracy.
\begin{table}[t]
    \caption{Ablation study on confidence threshold $\delta$ was conducted on CIFAR10. The green indicates the margins ahead SimGCD (\ie $\delta$=0), while the red donates lagging values.}
    \vspace{-3mm}
    \small
    \centering
    \resizebox{0.8\linewidth}{!}{
    \begin{tabular}{cccc}
    \toprule
        & \multicolumn{3}{c}{CIFAR10} \\ \cmidrule(lr){2-4}
        $\delta$ &All  &Old &New  \\ \hline
        0.0 &96.9 &93.8 &98.5  \\
        0.85 &97.5\scriptsize {\color{resultgreen}\textbf{+0.6}}  &96.4\scriptsize {\color{resultgreen}\textbf{+2.6}}  &98.0\scriptsize {\color{red}\textbf{-0.5}}  \\
        0.90 &97.4\scriptsize {\color{resultgreen}\textbf{+0.5}}   &96.0\scriptsize {\color{resultgreen}\textbf{+2.2}} 	&98.1\scriptsize {\color{red}\textbf{-0.4}}  \\
        0.95 &97.1\scriptsize {\color{resultgreen}\textbf{+0.2}}  &94.3\scriptsize {\color{resultgreen}\textbf{+0.5}}  &98.5\scriptsize {\color{resultgreen}\textbf{+0.0}}  \\
        0.97 &97.5\scriptsize {\color{resultgreen}\textbf{+0.6}}  &95.7\scriptsize {\color{resultgreen}\textbf{+1.9}}	&98.2\scriptsize {\color{red}\textbf{-0.3}}  \\
    \bottomrule
    \end{tabular}}
    \label{tab:cifar10_thr}
\end{table}

\begin{table}[t]
    \caption{The accuracy of the FGVC-Aircraft dataset compared with SimGCD and LegoGCD in different settings.}
    \vspace{-3mm}
    \small
    \centering
    \resizebox{\linewidth}{!}{
    \begin{tabular}{ccccccc}
    \toprule
        & \multicolumn{3}{c}{SimGCD} &\multicolumn{3}{c}{LegoGCD}\\ \cmidrule(lr){2-4} \cmidrule(lr){5-7}
        Seed &All  &Old &New   &All  &Old &New  \\ \hline
        Yes &54.6	&61.4  &51.1 & 55.0  &61.5\scriptsize {\color{resultgreen}\textbf{+0.1}} &51.7\scriptsize {\color{resultgreen}\textbf{+0.6}} \\ \hline
         &51.8 &57.2 &49.0 &53.5 &62.0	&49.2 \\
         &52.5 &58.3  &49.6 &54.6	&60.0	&51.9 \\
         &53.8  &58.8  &51.3 &56.1	&64.2	&52.0 \\
         &55.2 &61.8  &51.9 &55.8	&61.3	&53.0\\
          \multirow{-5}{*}{No}&56.6  &60.9  &54.4   &56.3	&62.7	&53.0\\ \hline
        Avg. &54.0\scriptsize {\color{resultgreen}\textbf{$\pm$1.75}}  &59.4\scriptsize {\color{resultgreen}\textbf{$\pm$1.67}}  &51.2\scriptsize {\color{resultgreen}\textbf{$\pm$1.94}}  &55.2\scriptsize {\color{resultgreen}\textbf{$\pm$1.18}}	&62.0\scriptsize {\color{resultgreen}\textbf{$\pm$1.57}}	&51.8\scriptsize {\color{resultgreen}\textbf{$\pm$1.57}} \\
    \bottomrule
    \end{tabular}}
    \label{tab:air}
\end{table}
\subsection{Results on FGVC-Aircraft}
In \cref{tab:air}, we analyze the accuracy in the FGVC-Aircraft dataset under different settings for comprehensive comparisons. Initially, we use the same random seed=0 in both SimGCD and LegoGCD. Subsequently, we conduct 5 training runs across SimGCD and LegoGCD without a fixed random seed and average the results. As depicted in \cref{tab:air}, when utilizing the same random seed=0, our method only slightly outperforms SimGCD by 0.1\%, as shown in \cref{tab:result_legegcd_1}. However, our method achieves a substantial improvement of 2.6\% in ``Old" classes after 5 runs. Additionally, the standard deviation of our method is 1.57 while 1.67 in SimGCD,  proving LegoGCD exhibits less fluctuation than SimGCD in the FGVC-Aircraft dataset.


\end{document}